\documentclass{article}
\usepackage{iclr2025_conference,times}

\usepackage[table]{xcolor}
\usepackage[T1]{fontenc}
\usepackage{hyperref}
\usepackage{url}
\usepackage{booktabs}
\usepackage{amsfonts}
\usepackage{nicefrac}
\usepackage{microtype}
\usepackage{amsmath}
\usepackage{amssymb}
\usepackage{amsthm}
\usepackage{graphicx}
\usepackage{framed}
\usepackage[thinc]{esdiff}

\usepackage{caption}
\usepackage{subcaption}
\usepackage{wrapfig}
\usepackage{multirow}
\usepackage{amsfonts}
\usepackage{bm}
\usepackage{verbatim}
\usepackage{longtable}
\usepackage{arydshln}
\usepackage{algorithm}
\usepackage{algpseudocode}
\usepackage{mathtools, nccmath}
\usepackage{pifont}

\usepackage{mathtools}
\usepackage{amsthm}
\usepackage{multirow}
\usepackage{graphicx}
\usepackage{lscape}
\usepackage{algorithm}
\usepackage{algpseudocode}
\usepackage[capitalize,noabbrev]{cleveref}

\usepackage[textsize=tiny]{todonotes}

\definecolor{mygreen}{HTML}{009901}
\definecolor{myred}{HTML}{A52A2A}

\newcommand{\cmark}{\ding{51}}
\newcommand{\xmark}{\ding{55}}

\newcommand{\pd}{\partial}
\DeclareDocumentCommand{\pdd}{ O{} O{} m }{\frac{\pd^{#2}\!#1}{\pd#3^{#2}}}

\DeclarePairedDelimiter{\nint}\lfloor\rceil

\DeclareMathOperator*{\argmin}{arg\,min}

\newcommand{\ours}{\texttt{SpinQuant}}
\newcommand{\ourse}{\texttt{SpinQuant$_{no\ \!had}$}}
\newcommand{\oursh}{\texttt{SpinQuant$_{had}$}}
\newcommand{\rotm}{\ensuremath{R}}
\newcommand{\hadm}{\ensuremath{H}}

\definecolor{mygreen}{HTML}{009901}
\definecolor{myred}{HTML}{A52A2A}
\iclrfinalcopy

\title{$\ours$: LLM Quantization with Learned Rotations}
\author{
\vspace{0.1em}
Zechun Liu\thanks{\hspace{.06in}Equal contribution. Correspondence to: Zechun Liu $<$zechunliu@meta.com$>$.}   \hspace{3mm}  Changsheng Zhao$^{*}$  \hspace{2mm} Igor Fedorov  \hspace{2mm} Bilge Soran \hspace{2mm} Dhruv Choudhary \\
\vspace{0.4em}
\textbf{Raghuraman Krishnamoorthi \hspace{2mm} Vikas Chandra  \hspace{2mm} Yuandong Tian \hspace{2mm} Tijmen Blankevoort} \\
\hspace{0.1mm} Meta \\
}
\begin{document}

\maketitle
    \begin{abstract}
    Post-training quantization (PTQ) techniques applied to weights, activations, and the KV cache greatly reduce memory usage, latency, and power consumption of Large Language Models (LLMs), but may lead to large quantization errors when outliers are present. Rotating activation or weight matrices helps remove outliers and benefits quantization. In this work, we identify a collection of applicable rotation parameterizations that lead to identical outputs in full-precision Transformer architectures while enhancing quantization accuracy. In addition, we find that some random rotations lead to much better quantization than others, with an up to \emph{13 points} difference in downstream zero-shot reasoning performance. As a result, we propose \ours{}, a novel approach that incorporates learned rotation matrices for optimal quantized network accuracy. With 4-bit quantization of weight, activation, and KV-cache, \ours{} narrows the accuracy gap on zero-shot reasoning tasks with full precision to merely 2.9 points on the LLaMA-2 7B model, surpassing LLM-QAT by 19.1 points and SmoothQuant by 25.0 points. Furthermore, SpinQuant also outperforms concurrent work QuaRot, which applies random rotations to remove outliers. In particular, for LLaMA-3 8B models that are hard to quantize, \ours{} reduces the gap to full precision by up to 45.1\% relative to QuaRot. Code is available at \href{https://github.com/facebookresearch/SpinQuant}{\texttt{github.com/facebookresearch/SpinQuant}}.
\end{abstract}

    \section{Introduction}
\label{sec:intro}

Large Language models (LLMs) have demonstrated impressive performance across many disciplines. SoTA open source models (\textit{e.g.}, LLaMA \citep{touvron2023llama}, Mistral \citep{jiang2023mistral}, etc) and proprietary LLMs (\textit{e.g.}, GPT \citep{achiam2023gpt}, Gemini\citep{team2023gemini}, etc) have been used in general purpose chatting assistants, medical diagnosticians \citep{thirunavukarasu2023large}, computer game content generators \citep{cox2023conversational}, coding co-pilots \citep{roziere2023code}, and much more. 

To serve such a high demand, the inference cost becomes a real issue. Many effective techniques have been developed. Post-training Quantization (PTQ), as one effective category of techniques, quantizes the weights (or activations) into low-precision and thus reduces the memory usage and may significantly improve latency. This is not only important for server-side inference, but also for on-device scenarios with small-sized LLMs \citep{liu2024mobilellm,llama3modelcard}. 

When applying quantization, outliers remain an open challenge because they stretch the quantization range, leaving fewer effective bits available for the majority of values. Prior research mitigates this challenge by trading quantization difficulty between weights and activations~\citep{xiao2022smoothquant,lin2023awq} or employing mixed-precision to handle outliers~\citep{zhao2023atom}. In this work, we focus on a new angle: multiplying the weight matrix with a rotation matrix to reduce outliers and enhance quantizability. Inspired by \citep{elhage2023privileged} and SliceGPT~\citep{ashkboos2023slicegpt}, we leverage the property of rotational invariance to construct rotation matrices in pairs from identity mapping, which can be integrated into nearby weights without affecting the overall network outputs. By applying these random rotations, we produce a distribution of weight or activation entries that is outlier-less, facilitating easy quantization.

In addition to using random rotation, which statistically works well, we find that the performance of quantized network could \emph{vary a lot} with different rotation matrices. For example, the downstream averaged accuracy on zero-shot reasoning tasks may change up to 13 points with different rotations. As a result, we propose \ours{} that \emph{integrates} and \emph{optimizes} the rotation matrix to minimize the final loss of the quantized network, with fixed weight parameters, by employing the \textit{Cayley SGD}~\citep{li2020efficient}, a proficient technique for optimizing orthonormal matrices. This optimization does not alter the full-precision network output but refines the intermediate activations and weights, making them more quantization-friendly.

In \ours{}, we introduce two rotation strategies tailored for different complexity levels: \ourse{} and \oursh{}. Here, \textit{had} refers to hadamard rotation matrix. In \ourse{}, as depicted in Figure~\ref{fig:overall_diagram}(b), we implement shortcut rotation ($\rotm_1$) and $W_v$-$W_o$ pair rotation ($\rotm_2$), which can be directly absorbed into the respective weight matrices. During inference, the original weights are simply replaced with the rotated quantized weights, eliminating the need for modification in the forward pass. Conversely, in \oursh{}, designed for scenarios with low-bit quantization of KV cache or activations (e.g., 4-bit), we further incorporate online Hadamard rotation matrices ($R_3, R_4$) to address activation outliers inside MLP block and KV cache.

To rigorously assess the effectiveness of \ours{}, we executed comprehensive experiments across seven leading Large Language Models (LLMs), including LLaMA-2\citep{touvron2023llama} models (7B/13B/70B), LLaMA-3\citep{llama3modelcard} models (1B/3B/8B), and the Mistral~\citep{jiang2023mistral} 7B model. The key contributions of this study are summarized as follows:

\begin{figure}[t!]
    \centering
    \includegraphics[width=\linewidth]{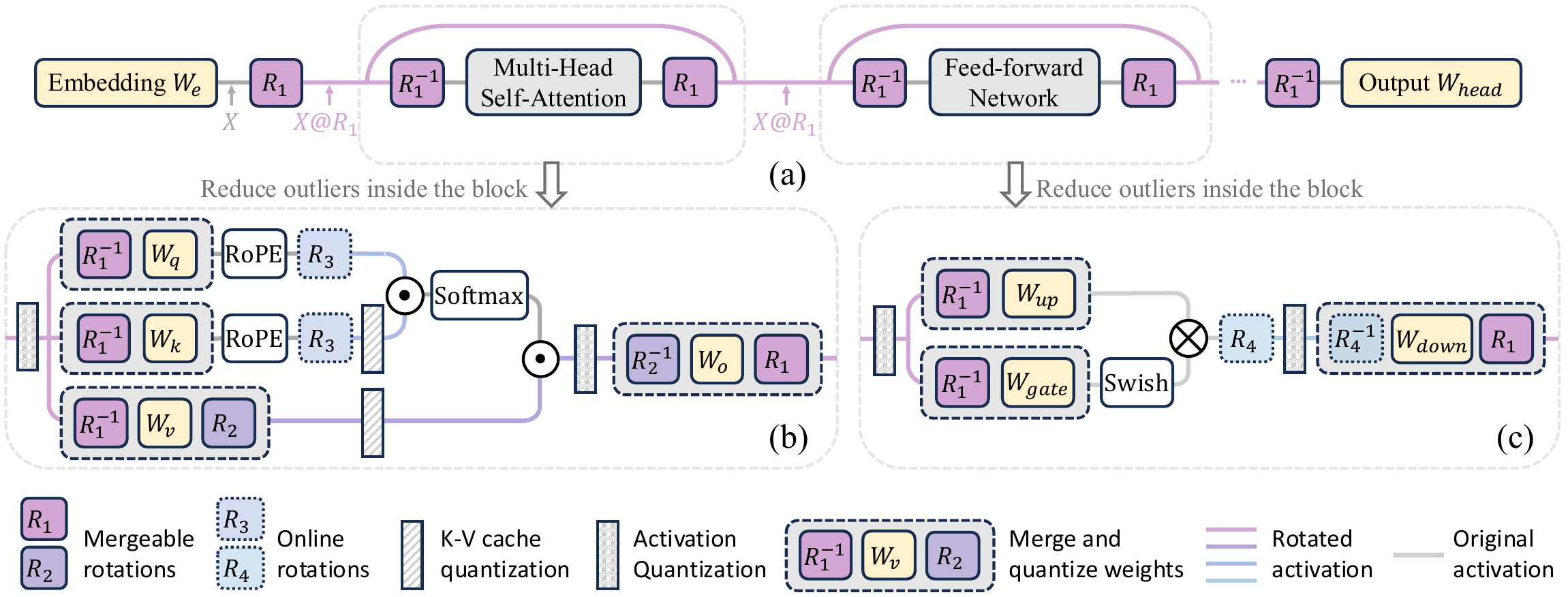}
    \caption{\small \textbf{Overall diagram of rotation.} (a) The residual stream can be rotated in the transformer network, resulting in numerically equivalent floating point networks before and after rotation. The rotated activations exhibit fewer outliers and are easier to quantize. (b) \& (c) The rotation matrix can be integrated with the corresponding weight matrices and we further define $R_2$, $R_3$, and $R_4$ for reducing outliers inside the block.} 
    \label{fig:overall_diagram}
\end{figure}

\begin{itemize}
\item We introduce \ours{}, the first method that employs learned rotations to mitigate outliers in weight and activation distributions, boosting the performance of quantized LLMs.
\item We reveal that random rotations introduce substantial variance in quantized network performance. We propose optimizing rotation matrices within \textit{Stiefel} manifold, directly minimizing the final loss of rotated quantized network. Ablation studies validate that our learned rotations consistently outperform random rotations, with improvements up to 16.2 points.
\item \ourse{} merges rotation matrices into pre-trained weights without altering the network architecture, significantly narrowing the W4A8KV8 quantization performance gap from 12.1 to 1.6 on the Mistral-7B model in zero-shot commonsense reasoning tasks. Noteworthily, \ourse{} W4A8 quantization achieves comparable performance as state-of-the-art weight only quantization methods like QuIP\#~\citep{tseng2024quip} and OminiQuant~\citep{shao2023omniquant} on LLaMA-2.
\item \oursh{} attains an average accuracy of 64.0 in extreme W4A4KV4 quantization settings on LLaMA-2 7B. This represents a mere 2.9 point gap from the full-precision network, a substantial improvement over the previous LLM-QAT~\citep{liu2023llmqat} approach, which exhibited a 22.0 point gap under identical precision conditions.
\end{itemize}
    \section{Motivation}
\begin{figure}[t!]
    \centering
    \includegraphics[width=0.9\linewidth]{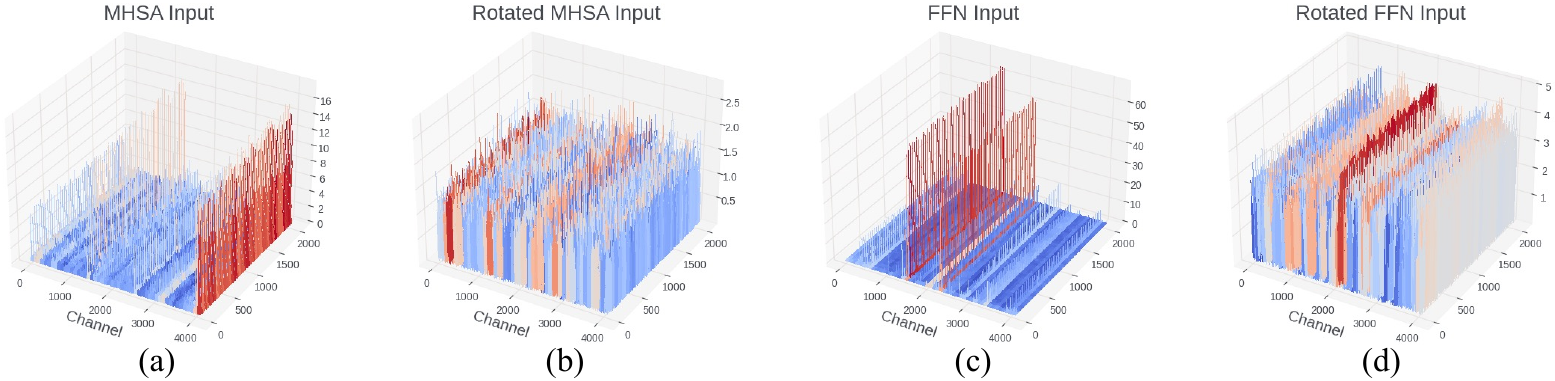}
    \caption{\small Activation distribution in LLaMA-2 7B model before and after rotation. Outliers exist in particular channels before rotation. Since channel-wise quantization is not supported in most hardware, outlier removal using rotation enables accurate token-wise or tensor-wise quantization.}  
    \label{fig:a_dist}
\end{figure}

\begin{figure}[t!]
    \centering
    \includegraphics[width=\linewidth]{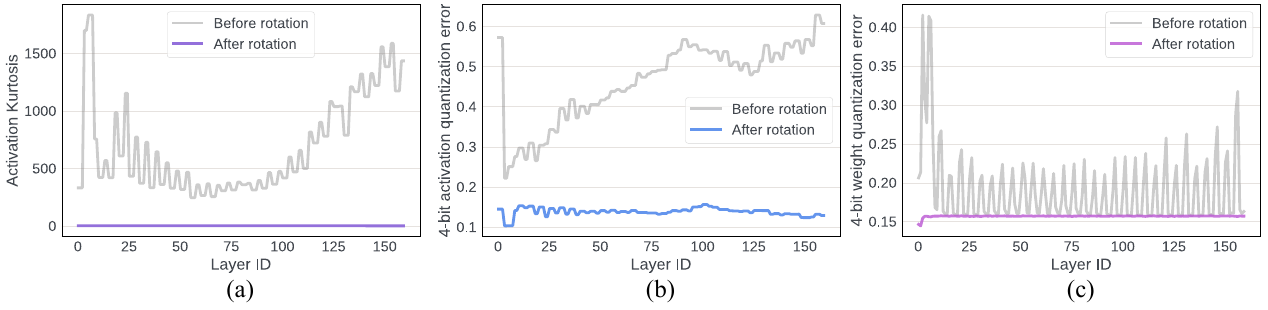}
    \caption{\small Outlier measurement and quantization error across input activation and weights in the five layers that take inputs from the residual (Q/K/V/Up/Gate-projection) of each block in the LLaMA-2 7B model. (a) After rotation, \textit{kurtosis} of activation distributions is significantly reduced to approximately three across all layers. Quantization error is reduced after rotation in both (b) activations and (c) weights.}
    \label{fig:kurtosis}
\end{figure}

Quantization reduces the precision of weights (and/or activations) in a neural network in order to save memory and lower the latency. The quantization process can be formulated as:
\begin{equation}
\label{eq:minmax}
    X_Q = \alpha \nint{\frac{X_R - \beta}{\alpha}} + \beta \ 
\end{equation}
where $\alpha = \frac{\max(|X_R|)}{2^{N-1} -1}, \beta = 0$ in symmetric quantization or $\alpha = \frac{\max(X_R) - \min(X_R)}{2^N -1}, \beta = \min(X_R)$ in asymmetric quantization. Here $X_Q$ is a quantized tensor and $X_R$ is a real-valued FP16 tensor. $N$ is number of bits.
For Large language models (LLMs), the presence of outliers extends the range of weight/activation values and increases the reconstruction errors for normal values~\citep{dettmers2022llmint8,liu2023llmfp4,bondarenko2023outliers} (Figures~\ref{fig:a_dist} (a)\&(c)).

\subsection{Outlier Reduction} 
There exist many ways to mitigate the effect of outliers~\citep{xiao2022smoothquant,dettmers2022llmint8}. In this paper, we propose to use optimized rotation to reduce outliers. Intuitively, a random rotation matrix statistically blends large and small weights together into a well-behaved distribution with fewer outliers~\citep{elhage2023privileged}, and thus is easier to quantize. 

Figure~\ref{fig:kurtosis} (a) illustrates the measurement of the \textit{Kurtosis} $\kappa$ of the activations before and after rotation. $\kappa$ quantifies the ``\textit{tailedness}'' of a real-valued random variable's probability distribution. A larger $\kappa$ indicates more outliers, while $\kappa \approx 3$ suggests a Gaussian-like distribution.  
In Figure~\ref{fig:kurtosis} (a), the activation distribution in the transformer contains numerous outliers, with $\kappa$ of many layers exceeding 200. However, after multiplying these activations with a random rotation matrix, the $\kappa$ across all layers becomes approximately 3, indicating a more Gaussian-shaped distribution that is easier to quantize. This is corroborated by Figure~\ref{fig:kurtosis} (b), where the quantization error of the activation tensor significantly decreases after rotation.

\begin{figure}[t!]
    \centering
    \includegraphics[width=0.6\linewidth]{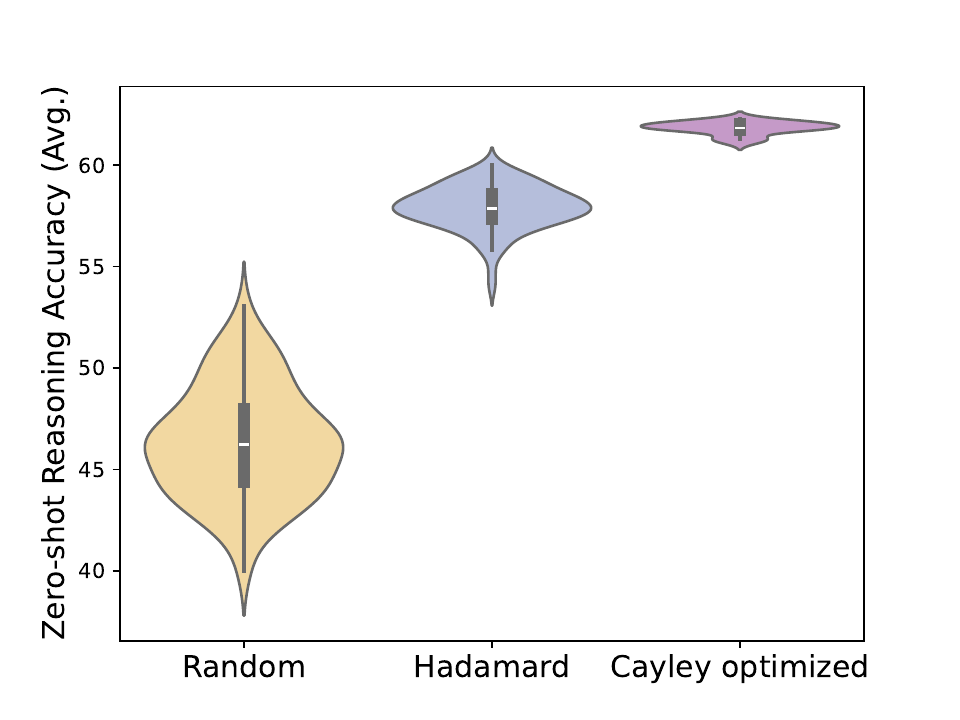}
    \caption{\small The performance distributions of W4A4 quantized LLaMA-2 7B under different random rotations, using network-level parameterization (Sec.~\ref{sec:network_level_rotation}). We compare the distributions using random floating-point rotations, random Hadamard matrices, and optimized rotation matrices with \textit{Cayley} optimization (Sec.~\ref{sec:cayley-optimized_rotation}). Despite that Hadamard matrices mostly perform better than random rotations, both random groups demonstrate large variance. In contrast, by optimizing the rotation matrix with \textit{Cayley} optimization (\textit{i.e.}, \ours{}), the performance is improved significantly and the variance becomes much smaller.}
    \label{fig:variance_rotation}
\end{figure}

\subsection{Random rotations produce large variance} 
\label{sec:random_rotation_variance}
Interestingly, while statistically random rotation leads to better quantization, not all random rotations give the same quantization outcome. To show this, we tested the zero-shot average accuracy of the rotated version of LLaMA-2 7B, quantized to 4-bit weight and 4-bit activation, under 100 randomized trials. As shown in Figure~\ref{fig:variance_rotation}, the performance variance is substantial, with the best random rotation matrix outperforming the worst by 13 points. Random Hadamard matrices~\footnote{A Hadamard matrix $\hadm$ is a special type of rotation matrix, where the entries of the matrix are solely $\pm \sqrt{n}$. Given a Hadamard matrix $\hadm$, we can generate $2^n$ different random Hadamard matrices by multiplying with $S$, a diagonal matrix with elements $s_i$ randomly chosen from $\{-1, 1 \}$.} outperform random rotation matrices, in consistent with the findings in~\citep{tseng2024quip} that Hadamard matrices yield tighter bounds on weight maximal value. However, even random Hadamard rotation matrices exhibit a non-negligible variance in final performance, as large as 6 points. 

Given the huge variance across multiple trials of rotations, a natural question arises: \emph{Is it possible to optimize the rotation to maximize the benefit of quantization?}
We affirmatively answer this question by presenting a viable framework with quantization-oriented rotation learning that consistently achieves high accuracy across 7 models and 4 low-bit quantization settings.

\section{Method}
In this section, we introduce \ours{}, a framework that integrates and optimizes rotations in LLMs targeting at quantization loss. We start with defining rotation parameterization of popular LLM architectures, which includes two mergeable rotation matrices ($\rotm_1$, $\rotm_2$) that produce rotationally invariant full-precision network, and two online Hadamard rotation ($\rotm_3$, $\rotm_4$) to further reduce the outliers for extreme activation and KV-cache quantization. Then, we present how to optimize these rotation matrices on \textit{Stiefel} manifold with target loss. 

\subsection{Rotation parameterization}
\label{sec:network_level_rotation}

\textbf{Rotating activations in residual}
As shown in Figure~\ref{fig:overall_diagram}(a), we rotate the activations in the residual path by multiplying the embedding output $X$ with a random rotation matrix ($\rotm_1$). This rotation removes outliers and eases the quantization of the input activations to the fully-connected layers that read from the residual. To maintain numerical invariance, we reverse the rotation of the activation by multiplying it with $\rotm_1^T$ ($=\rotm_1^{-1}$) prior to its passage through the attention block and feed-forward network, which contains non-linearity. When the quantization is not present, the full-precision network remains intact no matter which rotation is applied.\footnote{In a pre-norm LLM like LLaMA~\citep{llama}, we can convert a transformer network into a rotation-invariant network by incorporating the RMSNorm scale parameters $\alpha$ into the weight matrix right after the RMSNorm layer~\citep{ashkboos2023slicegpt}.} 
The rotation matrices can be merged into corresponding weight matrices, as illustrated in Figures~\ref{fig:overall_diagram}(b)\&(c). After absorption, no new parameters are introduced in the network. We can now modify $\rotm_1$ freely without impacting the floating-point network's accuracy or parameter count. 

\begin{figure}[h]
    \centering
    \includegraphics[width=0.5\linewidth]{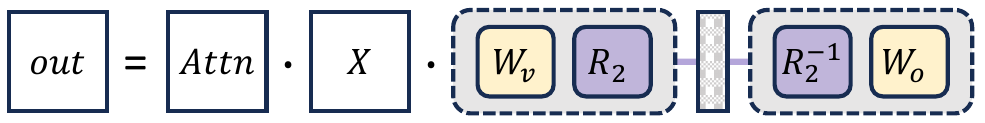}
    \vspace{-0.5em}
    \caption{\small Rotation equivalence in Multi-Head Self-Attention.}
    \label{fig:MHSA_rotation}
    \vspace{-1em}
\end{figure}

\textbf{Rotating activations in the attention block}
\label{sec:R2_optimization}
As depicted in Figure \ref{fig:overall_diagram}(b), in the attention block, we propose to rotate the value matrix by multiplying $\rotm_2$, and the activations to out-projection layer by $\rotm_2^T$ head-wisely. $\rotm_2$ has the shape of ($D_{head}, D_{head}$) and can be independently chosen across layers. The numerical in-variance is illustrated in Figure~\ref{fig:MHSA_rotation}, these two rotations can be offset in a full-precision network since there are no operators between $\rotm_2$ and $\rotm_2^T$. Meanwhile, it can improve quantization for value cache and input activations to out-projection layer without introducing any new parameters in the network. 

We denote the method with only $R_1$ and $R_2$ inserted and optimized as $\ourse{}$, which can readily achieve significant accuracy improvement than previous quantization methods, and closing the gap between W4A8 quantized LLMs and their full-precision counterparts to $0.1-2.5$ points on zero-shot commonsense reasoning averaged accuracy.

\textbf{Additional unabsorbed rotations}
To further enhance outlier suppression for lower-bit (\textit{e.g.} 4-bit) activation quantization, we incorporate a Hadamard matrix multiplication ($R_4$ in Figure~\ref{fig:overall_diagram}(c)) inside the feed-forward block, reducing the outliers in the input to the down projection layer, similar to~\citep{tseng2024quip,ashkboos2024quarot}. Hadamard rotation can be computed with fast hadamard transform and introduce marginal overhead to the inference latency. Similarly, Hadamard matrix ($R_3$ in Figure~\ref{fig:overall_diagram}(b)) can be inserted when low-bit KV cache quantization is required. We denote the resulting method, equipped with all rotations, as $\oursh{}$. Next, we demonstrate how to jointly optimize these rotations.

\subsection{\textit{Cayley}-optimized rotation}
\label{sec:cayley-optimized_rotation}

As illustrated in Figure \ref{fig:overall_diagram}, we have determined that the incorporation of four rotation matrices ($R_1$, $R_2$, $R_3$, $R_4$) can improve quantization performance while preserving numerical consistency in a full-precision network. Given that $R_3$ and $R_4$ are online rotation operations, meaning they cannot be absorbed into the weight matrix, we retain them as Hadamard matrices. This is because online Hadamard transforms can be efficiently implemented without significant overhead. We then define the optimization objective as identifying the optimal rotation matrix $R_1$ and $R_2$ that minimizes the final loss of the quantized network:
\begin{align}\label{eq:optimization_problem}
    \argmin_{\rotm \in \mathcal{M}} \mathcal{L}_Q(\rotm_1, \rotm_2 \mid W, X)
\end{align}
Here, $\mathcal{M}$ represents the \textit{Stiefel} manifold \textit{i.e.}, the set of all orthonormal matrices. $\mathcal{L}_Q(\cdot)$ denotes the task loss, such as cross-entropy, on the calibration set. It is a function of \{$\rotm_1$, $\rotm_2$\}, given the fixed pretrained weights $W$ and the input tensor $X$ and with the quantization function $Q$ in the network. To optimize the rotation matrix on the \textit{Stiefel} manifold, we employ the \textit{Cayley SGD} method \citep{li2020efficient}, which is an efficient optimization algorithm on the \textit{Stiefel} manifold. More specifically, in each iteration, the update of the rotation $R$ is parameterized as the following:
\begin{equation}
    R' = \Delta R(Y) R := \left(I - \frac{\alpha}{2}Y\right)^{-1} \left(I + \frac{\alpha}{2}Y \right) R \label{eq:r-update}
\end{equation}
where $\Delta R(Y):= (I - \frac{\alpha}{2}Y)^{-1} (I + \frac{\alpha}{2}Y )$ is the \emph{Cayley Transform} of a skew-symmetric matrix $Y$ (\textit{i.e.}, $Y^\top = -Y$). $Y$ is computed from a projection $\hat G$ of the gradient $G := \nabla_R \mathcal{L}_Q$ of the loss function: 
\begin{equation}
    Y = \hat G - \hat G^\top, \quad\quad \hat G := G R^\top - \frac12 R R^\top G R^\top 
\end{equation}
It can be shown that $\Delta R(Y)$ is always orthonormal and thus $R'$ is guaranteed to be orthonormal ($R'^\top R' = I$) if $R$ is orthonormal. While Eqn.~\ref{eq:r-update} requires a matrix inverse, the new rotation matrix $R'$ can be computed via an efficient fixed point iteration~\citep{li2020efficient}. Overall, the approach maintains the property of orthonormality with only $\sim$2 times the computation time per iteration compared to a naive SGD algorithm. 

We apply the \textit{Cayley SGD} method to solve Eqn.~\ref{eq:optimization_problem} for $\{R_1,R_2\}$, while the underlying weight parameters in the network remain frozen. $\{R_1,R_2\}$ count for only $\sim$0.26\% of the weight size and is constrained to be orthonormal. Consequently, the underlying floating-point network remains unchanged, and the rotation only influences the quantization performance.

By employing \textit{Cayley} optimization to update the rotation for 100 iterations on an 800-sample WikiText2 calibration dataset, we obtain a rotation matrix that outperforms the best random matrix and random Hadamard matrix in 100 random seeds, shown in Figure~\ref{fig:variance_rotation}. The \textit{Cayley}-optimized rotation exhibits minimal variance when initiated from different random seeds. The rotation matrices are initialized with random Hadamard matrices for optimization and our ablation study in Section~\ref{sec:rotation_type} demonstrates that the optimized rotation is robust to random rotation initialization as well.

    \section{Experiments}
\label{sec:experiments}
We conduct experiments on the LLaMA-2~\citep{touvron2023llama} models (7B/13B/70B), LLaMA-3~\citep{llama3modelcard} models (1B/3B/8B) and Mistral~\citep{jiang2023mistral} 7B model. Our evaluation of the proposed $\ours$ was carried out on eight zero-shot commonsense reasoning tasks. These tasks include BoolQ~\citep{clark2019boolq}, PIQA~\citep{bisk2020piqa}, SIQA~\citep{sap2019siqa}, HellaSwag~\citep{zellers2019hellaswag}, WinoGrande~\citep{sakaguchi2021winogrande}, ARC-easy and ARC-challenge~\citep{clark2018arc}, and OBQA~\citep{mihaylov2018obqa}. Additionally, we also report the perplexity score on WikiText2 testset~\citep{merity2016wiki2} for our evaluation.

\subsection{Experimental settings}
We employ \textit{Cayley SGD}~\citep{li2020efficient} to optimize the rotation matrix, $R_1$ and $R_2$, both initialized as a random Hadamard matrix, while maintaining all network weights constant. $R_1$ is the residual rotation, shaped as ($D_{token}, D_{token}$). $R_2$ is head-wise rotation in each attention block, shaped as ($D_{head}, D_{head}$) and is separately learned in each layer. The learning rate starts at 1.5 and linearly decays to 0. We utilize 800 samples from WikiText-2 to optimize rotation for 100 iterations. It takes only $\sim$ 13 / 18 / 30 minutes for LLaMA-3 1B / 3B / 8B, respectively, and $\sim$ 25 / 30 minutes for LLaMA-2 7B / 13B, respectively. For LLaMA-2 70B, it takes $\sim$ 3.5 hours and for Mistral-7B it takes $\sim$ 16 minutes. 

In the main results, we optimize the rotation with respect to the activation quantized network, where the weights remain 16-bit. After rotation is learned, we apply GPTQ on the rotated weights \citep{frantar2022gptq}, for which we adhere to the standard GPTQ settings by using 128 samples from WikiText-2 with a sequence length of 2048 as the calibration set for GPTQ quantization. In the main table, we present the results of $\ours{}$ with GPTQ, and in the ablation study, while we also show the results of employing simple round-to-nearest (RTN) quantization in the ablation study.

\subsection{Main results}
\begin{table}[btp]
\renewcommand\arraystretch{1}
\centering
\caption{\small Comparison of the perplexity score on WikiText2 and averaged accuracy on eight Zero-shot Common Sense Reasoning tasks. Results for SmoothQuant~\citep{xiao2022smoothquant}, LLM-QAT~\citep{liu2023llmqat}, GPTQ~\citep{frantar2022gptq} were obtained using their publicly released codebase. While OmniQuant~\citep{shao2023omniquant}, AWQ~\citep{lin2023awq}, and QuIP\#~\citep{tseng2024quip} results were quoted from their papers. Full results are in the Appendix. }
\vspace{-1.8em}
\label{tab:main}
\setlength{\tabcolsep}{0.8mm}
{\resizebox{\textwidth}{!}{
\begin{tabular}{c|l||cc:cc:cc||cc:cc:cc||cc}
& & & & & & & & & & & & & & & \\
\noalign{\vspace{0.1em}}\hline\noalign{\vspace{0.1em}}
\hline\noalign{\vspace{0.1em}}
 &  & \multicolumn{2}{c:}{\textbf{LLaMA-2 7B}} & \multicolumn{2}{c:}{\textbf{LLaMA-2 13B}} & \multicolumn{2}{c||}{\textbf{LLaMA-2 70B}} & \multicolumn{2}{c:}{\textbf{LLaMA-3.2 1B}} & \multicolumn{2}{c:}{\textbf{LLaMA-3.2 3B}} & \multicolumn{2}{c||}{\textbf{LLaMA-3 8B}} & \multicolumn{2}{c}{\textbf{Mistral-7B}} \\
\noalign{\vspace{0.1em}}\cdashline{3-15}\noalign{\vspace{0.1em}} 
\textbf{\#Bits} & \textbf{Method} & 0-shot$^8$ & Wiki & 0-shot$^8$ & Wiki & 0-shot$^8$ & Wiki & 0-shot$^8$ & Wiki & 0-shot$^8$ & Wiki & 0-shot$^8$ & Wiki & 0-shot$^8$ & Wiki \\
$_\text{(W-A-KV)}$ &  & Avg.($\uparrow$) & ($\downarrow$) & Avg.($\uparrow$) & ($\downarrow$) & Avg.($\uparrow$) & ($\downarrow$) & Avg.($\uparrow$) & ($\downarrow$) & Avg.($\uparrow$) & ($\downarrow$) & Avg.($\uparrow$) & ($\downarrow$) & Avg.($\uparrow$) & ($\downarrow$) \\
\noalign{\vspace{0.1em}}\hline\noalign{\vspace{0.1em}} 
\!\!16-16-16\! & FloatingPoint & 66.9 & 5.5 & 68.3 & 5.0 & 72.9 & 3.3 & 56.9 & 13.4 & 63.9 & 10.7 & 69.6 & 6.1 & 71.0 & 5.4 \\
\hdashline\noalign{\vspace{0.1em}} 
\multirow{9}{*}{4-8-16} & RTN & 62.4 & 7.9 & 57.3 & 6.7 & 68.6 & 5.0 & 55.4 & 20.7 & 58.6 & 29.0 & 65.5 & 8.2 & 59.3 & 6.8 \\
 & SmoothQuant & 58.9 & 7.5 & 63.6 & 6.1 & 70.6 & 4.1 & 47.1 & 1e2 & 55.6 & 3e2 & 61.0 & 10.7 & -- & -- \\
 & LLM-QAT & 64.8 & 11.4 & 67.5 & 14.5 & -- & -- & 53.2 & 21.0 & 60.8 & 41.1 & 67.2 & 7.7 & -- & -- \\
 & AWQ (w4) & -- & 6.2 & -- & 5.1 & -- & -- & -- & -- & -- & -- & -- & -- & -- & -- \\
 & OmniQuant (w4) & -- & 5.7 & -- & 5.0 & -- & 3.5 & -- & -- & -- & -- & -- & -- & -- & -- \\
 & QuIP\# (w4) & -- & \textbf{5.6} & -- & 5.0 & -- & \textbf{3.4} & -- & -- & -- & -- & -- & -- & -- & -- \\
 & GPTQ & 64.9 & 20.2 & 65.2 & 5.9 & 71.7 & 4.3 & 55.0 & 17.3 & 58.7 & 25.2 & 64.5 & 7.2 & 51.7 & 8.6 \\
\rowcolor{gray!20} \cellcolor{white} & $\ourse{}$ & \textbf{65.7} & 5.8 & \textbf{68.2} & 5.1 & 72.1 & 3.7 & 56.0 & 15.3 & 61.4 & 11.6 & \textbf{68.6} & 6.7 & 68.8 & 5.7 \\
\rowcolor{gray!20} \cellcolor{white} & $\oursh{}$ & \textbf{65.7} & 5.7 & 68.1 & \textbf{5.0} & \textbf{72.7} & 3.5 & \textbf{56.5} & \textbf{14.4} & \textbf{63.2} & \textbf{11.5} & 68.4 & \textbf{6.5} & \textbf{69.9} & \textbf{5.5} \\
\noalign{\vspace{0.1em}}\hdashline\noalign{\vspace{0.1em}} 
\multirow{6}{*}{4-8-8} & RTN & 62.5 & 7.9 & 57.6 & 6.7 & 68.4 & 5.0 & 55.7 & 20.7 & 58.4 & 28.8 & 65.3 & 8.2 & 58.9 & 6.7 \\
 & SmoothQuant & 58.8 & 7.5 & 63.4 & 6.1 & 70.5 & 4.1 & 47.1 & 1e2 & 55.5 & 3e2 & 60.9 & 10.7 & -- & -- \\
 & LLM-QAT & 64.6 & 11.4 & 67.5 & 14.2 & -- & -- & 53.1 & 21.0 & 60.5 & 39.3 & 66.9 & 7.6 & -- & -- \\
 & GPTQ & 64.8 & 20.2 & 65.3 & 5.9 & 71.6 & 4.3 & 54.8 & 17.3 & 58.7 & 24.1 & 64.6 & 7.2 & 51.7 & 8.6 \\
\rowcolor{gray!20} \cellcolor{white} & $\ourse{}$ & \textbf{65.8} & 5.8 & 68.1 & \textbf{5.1} & 72.2 & 3.7 & 55.7 & 15.3 & 61.8 & 11.7 & 68.6 & 6.7 & 69.4 & 5.7 \\
\rowcolor{gray!20} \cellcolor{white} & $\oursh{}$ & \textbf{65.8} & \textbf{5.7} & \textbf{68.2} & \textbf{5.1} & \textbf{72.7} & \textbf{3.5} & \textbf{55.8} & \textbf{14.3} & \textbf{63.2} & \textbf{11.2} & \textbf{68.8} & \textbf{6.5} & \textbf{70.2} & \textbf{5.5} \\
\noalign{\vspace{0.1em}}\hdashline\noalign{\vspace{0.1em}} 
\multirow{6}{*}{4-4-16} & RTN & 35.6 & 2e3 & 35.3 & 7e3 & 35.1 & 2e5 & 41.2 & 1e2 & 42.1 & 7e2 & 43.9 & 2e2 & 41.4 & 4e2 \\
 & SmoothQuant & 41.8 & 3e2 & 44.9 & 34.5 & 57.7 & 57.1 & 37.9 & 2e3 & 43.6 & 4e2 & 40.3 & 9e2 & -- & -- \\
 & LLM-QAT & 47.8 & 12.9 & 34.3 & 4e3 & -- & -- & 42.0 & 62.1 & 46.9 & 37.6 & 44.9 & 42.9 & -- & -- \\
 & GPTQ & 36.8 & 9e3 & 35.2 & 5e3 & 35.5 & 2e6 & 41.6 & 1e2 & 43.4 & 3e2 & 40.6 & 2e2 & 40.4 & 3e2 \\
\rowcolor{gray!20} \cellcolor{white} & $\ourse{}$ & 57.0 & 9.2 & 61.8 & 7.2 & 61.0 & 7.3 & 44.8 & 48.4 & 52.9 & 22.4 & 51.9 & 18.6 & 52.7 & 13.4 \\
\rowcolor{gray!20} \cellcolor{white} & $\oursh{}$ & \textbf{64.1} & \textbf{5.9} & \textbf{67.2} & \textbf{5.2} & \textbf{71.0} & \textbf{3.8} & \textbf{53.5} & \textbf{15.3} & \textbf{61.0} & \textbf{11.1} & \textbf{65.8} & \textbf{7.1} & \textbf{68.4} & \textbf{5.7} \\
\noalign{\vspace{0.1em}}\hdashline\noalign{\vspace{0.1em}} 
\multirow{6}{*}{4-4-4} & RTN & 37.1 & 2e3 & 35.5 & 7e3 & 35.0 & 2e5 & 40.6 & 2e2 & 41.2 & 8e2 & 43.1 & 3e2 & 41.4 & 4e2 \\
 & SmoothQuant & 39.0 & 7e2 & 40.5 & 56.6 & 55.9 & 10.5 & 36.5 & 2e3 & 40.0 & 6e2 & 38.7 & 2e3 & -- & -- \\
 & LLM-QAT & 44.9 & 14.9 & 35.0 & 4e3 & -- & -- & 41.5 & 76.2 & 45.9 & 42.0 & 43.2 & 52.5 & -- & -- \\
 & GPTQ & 36.8 & 9e3 & 35.2 & 5e3 & 35.6 & 1e6 & 41.6 & 1e2 & 41.1 & 4e2 & 40.5 & 2e2 & 41.3 & 2e2 \\
\rowcolor{gray!20} \cellcolor{white} & $\ourse{}$ & 56.0 & 9.2 & 60.7 & 7.1 & 62.0 & 7.4 & 45.3 & 47.7 & 52.9 & 22.4 & 52.6 & 18.6 & 52.4 & 13.7 \\
\rowcolor{gray!20} \cellcolor{white} & $\oursh{}$ & \textbf{64.0} & \textbf{5.9} & \textbf{66.9} & \textbf{5.3} & \textbf{71.2} & \textbf{3.8} & \textbf{53.4} & \textbf{15.9} & \textbf{60.5} & \textbf{11.4} & \textbf{65.5} & \textbf{7.3} & \textbf{68.6} & \textbf{5.8} \\
\noalign{\vspace{0.1em}}\hline\noalign{\vspace{0.1em}}
\hline
\end{tabular}}}
\vspace{-2em}
\end{table}

We present two rotation schemes $\ourse{}$ and $\oursh{}$ to accommodate different scenarios. In Table~\ref{tab:main}, we use seven models and four most commonly used bit-width settings to provide a guideline on which rotation scheme should be chosen in practice. 

Recap $\ourse{}$ uses learned rotation $\rotm_1$ and $\rotm_2$ only, which can be merged into corresponding model weights during inference time after the rotation is learned. Using $\ourse{}$ only needs to replace the original model weights with the rotated model weights, necessitating no modification to the forward pass nor any additional kernel support. While $\oursh{}$ comprises both learned rotations ($\rotm_1$, $\rotm_2$) and the online Hadamard rotations ($\rotm_3$, $\rotm_4$). During inference time, $\rotm_3$ and $\rotm_4$ can be computed with fast Hadamard kernel~\citep{tseng2024quip} and we show in Sec.~\ref{sec:speed}, the online Hadamard rotation only introduces $\sim$8\% of the network latency overhead.

As shown in Table~\ref{tab:main}, in the scenarios where weights are quantized to 4-bit and activations are quantized to 8-bit, using $\ourse{}$ can readily achieve good performance. For example, $\ourse{}$ enhances the 4-8-8 quantized Mistral 7B by 10.5 points. In llama3-8B, $\ourse{}$ achieves more than 4.1 point improvements compared to GPTQ~\citep{frantar2022gptq} on 4-8-16 setting, and leaving the gap to full-precision network to only 1.0 point. In these settings with activations not extremely quantized, using $\ourse{}$ is a viable solution, and adding additional online Hadamard rotation yields marginal benefit. 

In contrast, when activations are quantized to 4 bits, the accuracy drops significantly and most previous methods fail to produce meaningful results. $\ourse{}$ bridge the gap by up to 20 points. In 4-4-4 quantized LLaMA-2 models, $\ourse{}$ significantly surpasses LLM-QAT~\citep{liu2023llmqat}, by 11.1 points on 7B model and outperforms SmoothQuant~\citep{xiao2023smoothquant} by 20.2 on the 13B model, thereby reducing the gap to the corresponding full-precision network from 22.0 / 27.8 points to 10.9 / 7.6 points respectively. Still, the gap to the full-precision network is non-negligible. In this scenario, $\oursh{}$ can further improve the accuracy by more than 5 points and close the gap to the respective FP network to 2-4 points. In 4-4-4 quantized LLaMA-2 7B/13B/70B models, $\oursh{}$ leaves only a 2.9/1.4/1.7 accuracy gap to the corresponding full-precision network, significantly surpassing the previous SoTA methods by 19.1/16.4/15.3 points, respectively.

In addition, compared to the state-of-the-art weight-only quantization methods, OmniQuant~\citep{shao2023omniquant}, AWQ~\citep{lin2023awq} and QuIP\#~\citep{tseng2024quip}, $\ours{}$ achieves similar evaluation perplexity on Wiki dataset with 4-bit weights and 8-bit activations, and without using advance vector quantization technique. These results show $\ours{}$ is suitable for various scenarios and achieves state-of-the-art performance.

\subsection{Ablation studies}
\vspace{-0.5em}
\subsubsection{Learned rotation vs random rotation}
\begin{table}[btp]
\renewcommand\arraystretch{0.6}
\centering
\caption{\small Compared to Hadamard rotation, $\ours{}$ learned rotation consistently outperform by a significant margin. Results are averaged accuracy on eight Zero-shot CommonSense Reasoning tasks.} 
\label{tab:random_vs_learned}
\setlength{\tabcolsep}{1.5mm}
\vspace{-2em}
{\resizebox{0.82\textwidth}{!}{
\begin{tabular}{c|cc|cc|cc}
& & & & & & \\
& & & & & & \\
\noalign{\vspace{0.2em}}\hline\noalign{\vspace{0.1em}}
\hline\noalign{\vspace{0.2em}}
& \multicolumn{2}{c|}{\textbf{LLaMA-3.2 3B}} & \multicolumn{2}{c|}{\textbf{LLaMA-3 8B}} & \multicolumn{2}{c}{\textbf{Mistral-7B}} \\
\noalign{\vspace{0.2em}}
 & 4-4-16  &  4-4-4  &  4-4-16  &  4-4-4  &  4-4-16  &  4-4-4 \\
\noalign{\vspace{0.2em}}\hline\noalign{\vspace{0.1em}}
Random Hadamard $\rotm_{\{1,2\}}$ & 49.8  &  49.6  &  49.5  &  50.0  &  51.4  &  51.5 \\
$\ourse{}$ $\rotm_{\{1,2\}}$ & \textbf{52.9}\textcolor{mygreen}{$_{(\uparrow \mathbf{3.1})}$}  &  \textbf{52.9}\textcolor{mygreen}{$_{(\uparrow \mathbf{3.3})}$}  &  \textbf{51.9}\textcolor{mygreen}{$_{(\uparrow \mathbf{2.4})}$}  &  \textbf{52.6}\textcolor{mygreen}{$_{(\uparrow \mathbf{2.5})}$}  &  \textbf{52.7}\textcolor{mygreen}{$_{(\uparrow \mathbf{1.3})}$}  &  \textbf{52.4}\textcolor{mygreen}{$_{(\uparrow \mathbf{0.9})}$} \\
\noalign{\vspace{0.1em}}\hdashline\noalign{\vspace{0.1em}}
Random Hadamard $\rotm_{\{1,2,3,4\}}$ & 59.0  &  58.4  &  64.2  &  63.9  &  52.7  &  52.4 \\
$\oursh{}$ $\rotm_{\{1,2,3,4\}}$ & \textbf{61.0}\textcolor{mygreen}{$_{(\uparrow \mathbf{2.1})}$}  &  \textbf{60.5}\textcolor{mygreen}{$_{(\uparrow \mathbf{2.2})}$}  &  \textbf{65.8}\textcolor{mygreen}{$_{(\uparrow \mathbf{1.6})}$}  &  \textbf{65.5}\textcolor{mygreen}{$_{(\uparrow \mathbf{1.6})}$}  &  \textbf{68.4}\textcolor{mygreen}{$_{(\uparrow \mathbf{15.7})}$}  &  \textbf{68.6}\textcolor{mygreen}{$_{(\uparrow \mathbf{16.2})}$} \\
\noalign{\vspace{0.2em}}\hline\noalign{\vspace{0.1em}}
\hline
\end{tabular}}}
\vspace{-0.5em}
\end{table}

In Table~\ref{tab:random_vs_learned}, we contrast the use of random Hadamard rotations with $\ours{}$'s optimized rotations. Employing learned rotations, whether under $\rotm_{1,2}$ settings or $\rotm_{1,2,3,4}$ settings, consistently enhances accuracy across various models and bit-width configurations. Notably, in the quantization of Mistral-7B, \oursh{} secures an improvement exceeding 15.7 points over using random Hadamard rotations. Given that rotation optimization incurs a minimal time cost (only 30 minutes for smaller models and up to 3.5 hours for a 70B model) we advocate for the adoption of optimized rotations for precise quantization of LLMs.

\subsubsection{Compatibility with GPTQ}
\begin{table}[btp]
\renewcommand\arraystretch{0.6}
\centering
\caption{\small Ablation study on compatibility with GPTQ~\citep{frantar2022gptq} on a LLaMA2-7B model.} 
\label{tab:gptq_cayley}
\setlength{\tabcolsep}{1.5mm}
\vspace{-2em}
{\resizebox{0.55\textwidth}{!}{
\begin{tabular}{c|c|cc}
& & & \\
& & & \\
\noalign{\vspace{0.2em}}\hline\noalign{\vspace{0.1em}}
\hline\noalign{\vspace{0.2em}}
\textbf{\#Bits}$_{\text{(W-A-KV)}}$ & \textbf{Task} & \textbf{\textit{Cayley} on 4-4-KV} &\textbf{\textit{Cayley} on 16-4-KV} \\
\noalign{\vspace{0.2em}}\hline\noalign{\vspace{0.2em}}
\multirow{2}{*}{4-4-16} & 0-shot$^8$  Avg. & 61.0 $_{\pm 1.0}$ & 64.1 $_{\pm 0.4}$ \\
& Wiki & 6.7 $_{\pm 0.07}$ & 5.9 $_{\pm 0.00}$ \\
\noalign{\vspace{0.2em}}\hline\noalign{\vspace{0.2em}}
\multirow{2}{*}{4-4-4} & 0-shot$^8$  Avg. & 60.9 $_{\pm 0.6}$ & 64.0 $_{\pm 0.3}$ \\
& Wiki & 6.8 $_{\pm 0.15}$ & 5.9 $_{\pm 0.01}$ \\
\noalign{\vspace{0.2em}}\hline\noalign{\vspace{0.1em}}
\hline
\end{tabular}}}
\vspace{-1em}
\end{table}

In the context where both weights and activations are quantized, we observed that the learned rotations tend to adapt effectively to both weight and activation quantization. Given that GPTQ significantly helps mitigate the errors due to weight quantization, but leaves activation quantization untouched, we elect to optimize the rotation matrices with respect to a network where only activations are quantized. This approach allows the rotation to more efficiently manage the activation quantization error while leaving the weight quantization error to be addressed by GPTQ. As shown in Table~\ref{tab:gptq_cayley}, this modification resulted in superior performance in both W4A4 and W4A4KV4 settings in the LLaMA-2 7B model, which is the configuration we choose to utilize throughout the rest of this paper.

\subsubsection{Rotation type}
\label{sec:rotation_type}
\begin{table}[t]
\renewcommand\arraystretch{0.6}
\centering
\caption{\small Floating-point(FP) rotation vs Hadamard rotation on a LLaMA-2 7B model.}
\label{tab:had_vs_fp}
\setlength{\tabcolsep}{1.5mm}
\vspace{-2.5em}
{\resizebox{0.7\textwidth}{!}{
\begin{tabular}{c|c|cc|ccc}
& & & & & & \\
& & & & & & \\
& & & & & & \\
\noalign{\vspace{0.2em}}\hline\noalign{\vspace{0.1em}}
\hline\noalign{\vspace{0.2em}}
\textbf{\#Bits} & & \multicolumn{2}{c|}{No \textit{Cayley} + RTN} & \multicolumn{2}{c}{\textit{Cayley} + RTN} \\
\noalign{\vspace{0.1em}}
(W-A-KV) & Task  & FP & Hadamard  & FP init. & Hadamard init.\\
\noalign{\vspace{0.2em}}\hline\noalign{\vspace{0.2em}}
\multirow{2}{*}{4-16-16} & 0-shot$^8$  Avg.($\uparrow$) & 62.5 $_{\pm 0.8}$ & 62.4 $_{\pm 1.0}$ & 64.9 $_{\pm 0.4}$ & 64.6 $_{\pm 0.3}$ \\
& Wiki($\downarrow$) & 6.7 $_{\pm 0.12}$ & 6.9 $_{\pm 0.45}$ & 5.5 $_{\pm 0.01}$ & 5.5 $_{\pm 0.01}$ \\
\noalign{\vspace{0.2em}}\hline\noalign{\vspace{0.2em}}
\multirow{2}{*}{4-4-16} & 0-shot$^8$  Avg.($\uparrow$) & 49.4 $_{\pm 2.8}$ & 59.0 $_{\pm 1.0}$ & 61.6 $_{\pm 0.4}$ & 61.8 $_{\pm 0.4}$ \\
& Wiki($\downarrow$) & 15.9 $_{\pm 4.04}$ & 8.2 $_{\pm 0.73}$ & 6.2 $_{\pm 0.06}$ & 6.1 $_{\pm 0.03}$ \\
\noalign{\vspace{0.2em}}\hline\noalign{\vspace{0.2em}}
\multirow{2}{*}{4-4-4} & 0-shot$^8$  Avg.($\uparrow$) & 48.3 $_{\pm 2.7}$ & 58.7 $_{\pm 1.0}$ & 61.5 $_{\pm 0.8}$ & 61.5 $_{\pm 0.3}$ \\
& Wiki($\downarrow$) & 18.2 $_{\pm 4.35}$ & 8.2 $_{\pm 0.36}$ & 6.3 $_{\pm 0.08}$ & 6.2 $_{\pm 0.03}$ \\
\noalign{\vspace{0.2em}}\hline\noalign{\vspace{0.1em}}
\hline
\end{tabular}}}
\end{table}

In Table~\ref{tab:had_vs_fp}, we evaluate the impact of random orthogonal floating-point rotation matrices and random Hadamard matrices on quantization accuracy, utilizing round-to-nearest quantization for our analysis. Prior to optimization, the Hadamard matrices yield a better-quantized network performance compared to floating-point rotation matrices.
However, after optimization, the initial choice of rotation, whether floating-point or Hadamard, becomes less significant. This is likely due to the loss-aware rotation optimization's ability to locate an optimal local minima that effectively minimizes quantization error, thereby enhancing robustness to varying types of rotation initialization.

\subsubsection{Comparison with QuaRot}

\begin{table}[btp]
\renewcommand\arraystretch{0.6}
\centering
\caption{\small Comparison with QuaRot~\citep{ashkboos2024quarot}.} 
\label{tab:quarot}
\setlength{\tabcolsep}{1.5mm}
\vspace{-2.5em}
{\resizebox{0.75\textwidth}{!}{
\begin{tabular}{c||cc:cc||cc:cc}
& \multicolumn{4}{c||}{}  & \multicolumn{4}{c}{} \\
& \multicolumn{4}{c||}{}  & \multicolumn{4}{c}{} \\
& & & & & & \\
\noalign{\vspace{0.2em}}\hline\noalign{\vspace{0.1em}}
\hline\noalign{\vspace{0.2em}}
 & \multicolumn{4}{c||}{\cellcolor{white}\textbf{LLaMA-3 8B} (FP: 69.6, 6.1)} & \multicolumn{4}{c}{\cellcolor{white}\textbf{LLaMA-3 70B} (FP: 74.5, 2.8)} \\
\noalign{\vspace{0.2em}}
 & \multicolumn{2}{c:}{4-4-16} & \multicolumn{2}{c||}{4-4-4} & \multicolumn{2}{c:}{4-4-16} & \multicolumn{2}{c}{4-4-4} \\
\noalign{\vspace{0.2em}}\cline{2-9}\noalign{\vspace{0.1em}}
 & 0-shot$^8$ & Wiki & 0-shot$^8$ & Wiki & 0-shot$^8$ & Wiki & 0-shot$^8$ & Wiki \\
 & Avg.($\uparrow$) & ($\downarrow$) & Avg.($\uparrow$) & ($\downarrow$) & Avg.($\uparrow$) & ($\downarrow$) & Avg.($\uparrow$) & ($\downarrow$) \\
\noalign{\vspace{0.2em}}\hline\noalign{\vspace{0.1em}}
QuaRot+RTN     & 59.5 & 10.4 & 58.6 & 10.9 & 41.5 & 91.2 & 41.3 & 92.4 \\\noalign{\vspace{0.2em}}
$\oursh{}$+RTN & \textbf{64.6} & \textbf{7.7} & \textbf{64.1} & \textbf{7.8} & \textbf{70.1} & \textbf{4.1} & \textbf{70.1} & \textbf{4.1} \\
\noalign{\vspace{0.2em}}\hdashline\noalign{\vspace{0.2em}}
QuaRot+GPTQ & 63.8 & 7.9 & 63.3 & 8.0 & 65.4 & 20.4 & 65.1 & 20.2 \\\noalign{\vspace{0.2em}}
$\oursh{}$+GPTQ & \textbf{65.8} & \textbf{7.1} & \textbf{65.5} & \textbf{7.3} & \textbf{69.5} & \textbf{5.5} & \textbf{69.3} & \textbf{5.5} \\
\noalign{\vspace{0.2em}}\hline\noalign{\vspace{0.1em}}
\hline
\end{tabular}}}
\end{table}

Compared to QuaRot~\citep{ashkboos2024quarot}, which exhibits significant accuracy variances in quantized networks—experiencing drops of 28.1 and 33.2 points when quantizing a 70B model with round-to-nearest methods to W4A4 and W4A4KV4—this degradation stems from inherent noise in using random rotation matrix that introduce high variance and compromise robustness. In contrast, $\oursh{}$ consistently maintains high accuracy across various configurations, achieving improvements of 2.0 to 28.6 points over QuaRot (Table~\ref{tab:quarot}), while utilizing fewer online Hadamard matrices (two per block in $\oursh{}$ versus four per block in QuaRot).

Furthermore, the integration of $\rotm_2$ in \ours{} effectively reduces in-block outliers, thereby enabling \ourse{} to deliver optimal performance in W4A8 settings. \ourse{} can be achieved by simply substituting the model weights with rotated weights, making it a more straightforward and efficient approach compared to QuaRot, which requires modifying the model architecture and special kernel support.

\subsection{Illustrative analysis of the rotation efficacy}
\begin{figure}[b!]
    \centering
    \vspace{-1em}
    \includegraphics[width=0.85\linewidth]{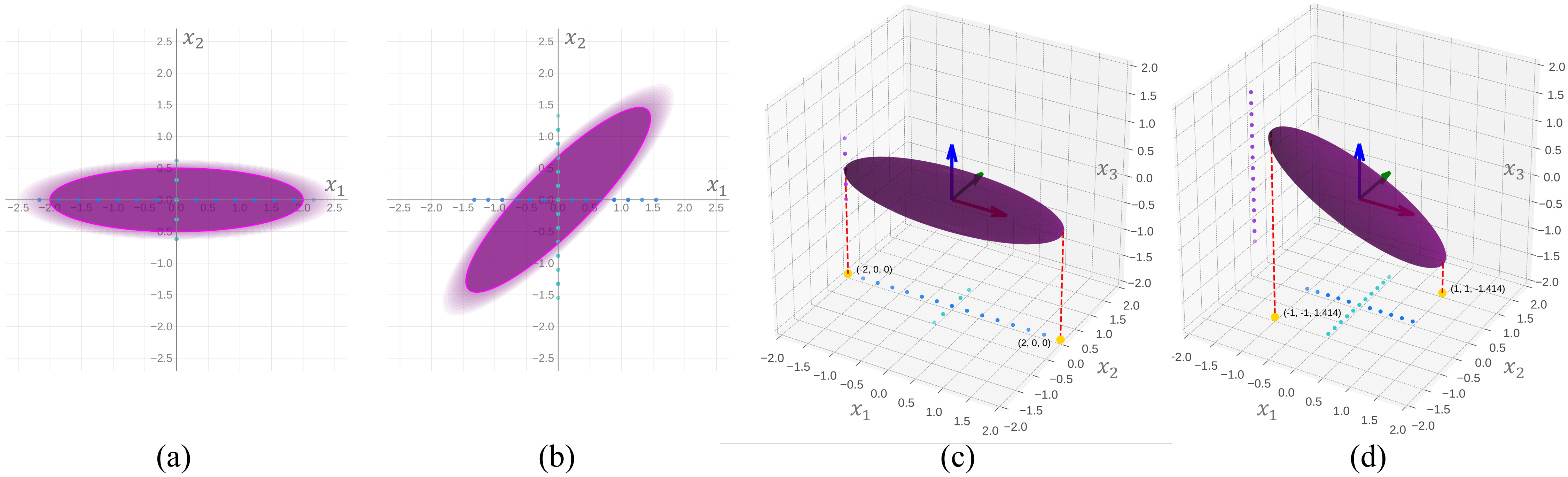}
    \vspace{-1em}
    \caption{\small An illustration of how rotation helps reduce outliers and maximize quantization range utilization.}
    \label{fig:understanding}
\end{figure}
The rationale behind rotating network weights and activations can be elucidated through a straightforward example. Consider an activation ($X$) represented as a 2D vector, where one entry $x_1$ consistently receives higher magnitude activations than $x_2$ (as depicted in Figure~\ref{fig:understanding}(a)). Quantizing these components together typically results in a quantization range dominated by $x_1$, thereby compromising the precision for $x_2$.

From an information entropy standpoint, expanding each axis to fully utilize the available quantization range maximizes the representational capacity of each axis. Thus, matrix rotation emerges as an intuitive solution. In a 2D scenario, rotating the axis by 45° equalizes the value representation range across axes (illustrated in Figure~\ref{fig:understanding}(b)). Assuming the network as a black box without knowledge of the exact activation distribution, uniformly rotating all axes by the maximal degree (45° in 2D) can optimize distribution evenness across each axis, partially explaining why Hadamard rotation often outperforms random rotation matrices.

Taking this further, if the activation distribution is known, treating the network as a white box during quantization allows for the identification of more optimal rotations than Hadamard. For instance, in a 3D scenario depicted in Figure~\ref{fig:understanding}(c-d), where $x_1$'s magnitude is four times that of $x_2$ and $x_3$, rotating the distribution by 45° along $x_3$ and $x_2$ redistributes the maximum values from $[2, 0.5, 0.5]$ to $[1, 1, 1.414]$. However, even more optimal rotation strategies may exist, and learning the rotation can help pinpoint the most effective rotation for a given distribution.

This opens up intriguing research avenues, such as determining if, given an activation distribution with known outlier axes and magnitudes, a closed-form solution for the optimal rotation matrix that evenly distributes magnitude across different axes can be derived. Additionally, it raises the question of whether this theoretically calculated rotation yields the best quantization performance. We leave this question to future research.
 
\subsection{Speed measurement}
\label{sec:speed}
\begin{table}[t]
\renewcommand\arraystretch{0.6}
\centering
\caption{\small Real-time end-to-end speed measurement of LLaMA-3 8B on MacBook M1 Pro CPU. }
\label{tab:speed}
\setlength{\tabcolsep}{1.5mm}
\vspace{-1em}
{\resizebox{0.45\textwidth}{!}{
\begin{tabular}{lcc}
\hline\noalign{\vspace{0.2em}}
\hline\noalign{\vspace{0.2em}}
\textbf{Method} & \#\textbf{Bits}$_{\text{(W-A)}}$ & \textbf{Decoding speed} \\
\hline\noalign{\vspace{0.2em}}
FloatingPoint & 16-16 & 177.15 ms/token \\
$\ourse{}$ & 4-8 & 58.88 ms/token \\
$\oursh{}$ & 4-8 &  63.90 ms/token \\
\hline\noalign{\vspace{0.2em}}
\hline\noalign{\vspace{0.2em}}
\end{tabular}}}
\vspace{-1em}
\end{table}

We conduct an end-to-end speed measurement of the LLaMA-3 8B model with W16A16 and W4A8 configurations on a MacBook M1 Pro CPU (OS version 14.5). The results in Table~\ref{tab:speed} demonstrate that 4-bit quantization yields a $\sim$3$\times$ speedup compared to the 16-bit model. Comparing $\oursh{}$ to $\ourse{}$, online Hadamard processing introduced a modest 8\% increase in latency. Therefore, it is a trade-off between using $\ourse{}$ without online Hadamard for its simpleness or using $\oursh{}$ with online Hadamard rotations for higher accuracy in lower-bit activation quantization. Detailed GPU latency results are provided in the Appendix.

    \section{Related Work}
\label{sec:related_work}

\textbf{Quantization}
Neural network quantization has been demonstrated as an effective tool for model size compression and storage reduction~\citep{adaround,krishnamoorthi,nagel2019data,brecq}. 
However, in large language models (LLMs), quantization presents unique challenges due to the presence of numerous outliers. These outliers dominate the quantization range, leaving only a few effective bits for the majority of values. Various strategies have been proposed to address the difficulties in LLM quantization. These include separating outliers and using mixed precision~\citep{dettmers2022llmint8,baalen2023gptvq,kim2023squeezellm,huang2024slim,egiazarian2024aqlm}, employing Hessian-based methods to mitigate quantization difficulty~\citep{frantar2022gptq}, trading outliers between weights and activations~\citep{xiao2022smoothquant,lin2023awq,liu2023llmfp4} utilizing weight equalization~\citep{nagel2019data}, outlier suppression~\citep{wei2022outlier,wei2023outlier},  channel reassembly~\citep{liu2023qllm} and even suggesting architectural modifications to handle outliers during pre-training\citep{bondarenko2023outliers}. Recently two QuIP papers~\citep{chee2024quip,tseng2024quip} introduce the incoherence processing using random rotation matrices and applying vector quantization on the weights for compression. This does introduce extra overhead and imposes some constraints on the devices the LLM is deployed to in the availability of vector quantization kernels.

\textbf{Optimization in orthonormal space}
The optimization of rotation matrices is carried out within the \textit{Stiefel Manifold}~\citep{james1976topology}, which encompasses all orthonormal matrices. Optimization while staying on this manifold can be done by \textit{e.g.}, parameterizing a skew-symmetric matrix and applying the \textit{Cayley} transformation on top of it \citep{nishimori2005learning}, or using a matrix exponential \citep{absil2012projection, casado2019cheaporthogonal}. However, these methods rely on expensive inverse or matrix-exponential functions that are applied every iteration. Instead, we follow the more efficient method named \textit{Cayley SGD}~\citep{li2020efficient}, which can be applied to optimize a rotation matrix $\rotm$ for arbitrary loss functions efficiently. \textit{Cayley SGD} relies on an iterative approximation of the \textit{Cayley} Transform that is conducted solely with matrix multiplications. 

    \section{Conclusions}
\label{sec:conclusions}
In this paper, we present \ours, a novel quantization technique that utilizes learned rotation to effectively bridge the performance gap between full precision and 4-bit weight, activation, and kv-cache quantization. At its core, \ours{} leverages the rotation invariance property of LLM models to insert rotation matrices that diminish outliers in the weights and intermediate activations while maintaining the network's full-precision output numerically identical. Additionally, \ours{} incorporates \textit{Cayley SGD} for optimizing rotation matrices, resulting in improved and robust quantization outcomes. Importantly, \ours{} is compatible with more advanced weight quantization techniques (\textit{e.g.}, GPTQ) and demonstrates state-of-the-art performance.

    \section* {Acknowledgment}
We extend our gratitude to Scott Wolchok and Chen Lai for their crucial contributions to latency measurement on the MacBook M1 Pro CPU, and to Sijia Chen, Geonhwa Jeong and Jiecao Yu for their expert support in GPU latency assessment.
    {\small
    \bibliography{references}

\begin{thebibliography}{50}
\providecommand{\natexlab}[1]{#1}
\providecommand{\url}[1]{\texttt{#1}}
\expandafter\ifx\csname urlstyle\endcsname\relax
  \providecommand{\doi}[1]{doi: #1}\else
  \providecommand{\doi}{doi: \begingroup \urlstyle{rm}\Url}\fi

\bibitem[Absil and Malick(2012)]{absil2012projection}
P-A Absil and J{\'e}r{\^o}me Malick.
\newblock Projection-like retractions on matrix manifolds.
\newblock \emph{SIAM Journal on Optimization}, 22\penalty0 (1):\penalty0 135--158, 2012.

\bibitem[Achiam et~al.(2023)Achiam, Adler, Agarwal, Ahmad, Akkaya, Aleman, Almeida, Altenschmidt, Altman, Anadkat, et~al.]{achiam2023gpt}
Josh Achiam, Steven Adler, Sandhini Agarwal, Lama Ahmad, Ilge Akkaya, Florencia~Leoni Aleman, Diogo Almeida, Janko Altenschmidt, Sam Altman, Shyamal Anadkat, et~al.
\newblock Gpt-4 technical report.
\newblock \emph{arXiv preprint arXiv:2303.08774}, 2023.

\bibitem[AI@Meta(2024)]{llama3modelcard}
AI@Meta.
\newblock Llama 3 model card.
\newblock 2024.
\newblock URL \url{https://github.com/meta-llama/llama3/blob/main/MODEL_CARD.md}.

\bibitem[Ashkboos et~al.(2023{\natexlab{a}})Ashkboos, Croci, do~Nascimento, Hoefler, and Hensman]{ashkboos2023slicegpt}
Saleh Ashkboos, Maximilian~L Croci, Marcelo~Gennari do~Nascimento, Torsten Hoefler, and James Hensman.
\newblock Slicegpt: Compress large language models by deleting rows and columns.
\newblock In \emph{The Twelfth International Conference on Learning Representations}, 2023{\natexlab{a}}.

\bibitem[Ashkboos et~al.(2023{\natexlab{b}})Ashkboos, Mohtashami, Croci, Li, Jaggi, Alistarh, Hoefler, and Hensman]{ashkboos2024quarot}
Saleh Ashkboos, Amirkeivan Mohtashami, Maximilian~L. Croci, Bo~Li, Martin Jaggi, Dan Alistarh, Torsten Hoefler, and James Hensman.
\newblock Quarot: Outlier-free 4-bit inference in rotated llms.
\newblock 2023{\natexlab{b}}.

\bibitem[Bisk et~al.(2020)Bisk, Zellers, Gao, Choi, et~al.]{bisk2020piqa}
Yonatan Bisk, Rowan Zellers, Jianfeng Gao, Yejin Choi, et~al.
\newblock Piqa: Reasoning about physical commonsense in natural language.
\newblock In \emph{Proceedings of the AAAI conference on artificial intelligence}, volume~34, pages 7432--7439, 2020.

\bibitem[Chee et~al.(2024)Chee, Cai, Kuleshov, and De~Sa]{chee2024quip}
Jerry Chee, Yaohui Cai, Volodymyr Kuleshov, and Christopher~M De~Sa.
\newblock Quip: 2-bit quantization of large language models with guarantees.
\newblock \emph{Advances in Neural Information Processing Systems}, 36, 2024.

\bibitem[Clark et~al.(2019)Clark, Lee, Chang, Kwiatkowski, Collins, and Toutanova]{clark2019boolq}
Christopher Clark, Kenton Lee, Ming-Wei Chang, Tom Kwiatkowski, Michael Collins, and Kristina Toutanova.
\newblock Boolq: Exploring the surprising difficulty of natural yes/no questions.
\newblock \emph{arXiv preprint arXiv:1905.10044}, 2019.

\bibitem[Clark et~al.(2018)Clark, Cowhey, Etzioni, Khot, Sabharwal, Schoenick, and Tafjord]{clark2018arc}
Peter Clark, Isaac Cowhey, Oren Etzioni, Tushar Khot, Ashish Sabharwal, Carissa Schoenick, and Oyvind Tafjord.
\newblock Think you have solved question answering? try arc, the ai2 reasoning challenge.
\newblock \emph{arXiv preprint arXiv:1803.05457}, 2018.

\bibitem[Cox and Ooi(2023)]{cox2023conversational}
Samuel~Rhys Cox and Wei~Tsang Ooi.
\newblock Conversational interactions with npcs in llm-driven gaming: Guidelines from a content analysis of player feedback.
\newblock In \emph{International Workshop on Chatbot Research and Design}, pages 167--184. Springer, 2023.

\bibitem[Dettmers et~al.(2022)Dettmers, Lewis, Belkada, and Zettlemoyer]{dettmers2022llmint8}
Tim Dettmers, Mike Lewis, Younes Belkada, and Luke Zettlemoyer.
\newblock Llm.int8(): 8-bit matrix multiplication for transformers at scale.
\newblock 2022.

\bibitem[Egiazarian et~al.(2024)Egiazarian, Panferov, Kuznedelev, Frantar, Babenko, and Alistarh]{egiazarian2024aqlm}
Vage Egiazarian, Andrei Panferov, Denis Kuznedelev, Elias Frantar, Artem Babenko, and Dan Alistarh.
\newblock Extreme compression of large language models via additive quantization.
\newblock \emph{arXiv preprint arXiv:2401.06118}, 2024.

\bibitem[Elhage et~al.(2023)Elhage, Lasenby, and Olah]{elhage2023privileged}
Nelson Elhage, Robert Lasenby, and Christopher Olah.
\newblock Privileged bases in the transformer residual stream.
\newblock \emph{Transformer Circuits Thread}, 2023.

\bibitem[Frantar et~al.(2022)Frantar, Ashkboos, Hoefler, and Alistarh]{frantar2022gptq}
Elias Frantar, Saleh Ashkboos, Torsten Hoefler, and Dan Alistarh.
\newblock Gptq: Accurate post-training quantization for generative pre-trained transformers.
\newblock \emph{arXiv preprint arXiv:2210.17323}, 2022.

\bibitem[Huang et~al.(2024)Huang, Qin, Liu, Li, Liu, Benini, Magno, and Qi]{huang2024slim}
Wei Huang, Haotong Qin, Yangdong Liu, Yawei Li, Xianglong Liu, Luca Benini, Michele Magno, and Xiaojuan Qi.
\newblock Slim-llm: Salience-driven mixed-precision quantization for large language models.
\newblock \emph{arXiv preprint arXiv:2405.14917}, 2024.

\bibitem[James(1976)]{james1976topology}
Ioan~Mackenzie James.
\newblock \emph{The topology of Stiefel manifolds}, volume~24.
\newblock Cambridge University Press, 1976.

\bibitem[Jiang et~al.(2023)Jiang, Sablayrolles, Mensch, Bamford, Chaplot, Casas, Bressand, Lengyel, Lample, Saulnier, et~al.]{jiang2023mistral}
Albert~Q Jiang, Alexandre Sablayrolles, Arthur Mensch, Chris Bamford, Devendra~Singh Chaplot, Diego de~las Casas, Florian Bressand, Gianna Lengyel, Guillaume Lample, Lucile Saulnier, et~al.
\newblock Mistral 7b.
\newblock \emph{arXiv preprint arXiv:2310.06825}, 2023.

\bibitem[Kim et~al.(2023)Kim, Hooper, Gholami, Dong, Li, Shen, Mahoney, and Keutzer]{kim2023squeezellm}
Sehoon Kim, Coleman Hooper, Amir Gholami, Zhen Dong, Xiuyu Li, Sheng Shen, Michael~W Mahoney, and Kurt Keutzer.
\newblock Squeezellm: Dense-and-sparse quantization.
\newblock \emph{arXiv preprint arXiv:2306.07629}, 2023.

\bibitem[{Krishnamoorthi}(2018)]{krishnamoorthi}
Raghuraman {Krishnamoorthi}.
\newblock {Quantizing deep convolutional networks for efficient inference: A whitepaper}.
\newblock \emph{arXiv preprint arXiv:1806.08342}, 2018.

\bibitem[Lezcano-Casado and Martinez-Rubio(2019)]{casado2019cheaporthogonal}
Mario Lezcano-Casado and David Martinez-Rubio.
\newblock Cheap orthogonal constraints in neural networks: A simple parametrization of the orthogonal and unitary group.
\newblock 2019.

\bibitem[Li et~al.(2020)Li, Fuxin, and Todorovic]{li2020efficient}
Jun Li, Li~Fuxin, and Sinisa Todorovic.
\newblock Efficient riemannian optimization on the stiefel manifold via the cayley transform.
\newblock \emph{arXiv preprint arXiv:2002.01113}, 2020.

\bibitem[Li et~al.(2021)Li, Gong, Tan, Yang, Hu, Zhang, Yu, Wang, and Gu]{brecq}
Yuhang Li, Ruihao Gong, Xu~Tan, Yang Yang, Peng Hu, Qi~Zhang, Fengwei Yu, Wei Wang, and Shi Gu.
\newblock Brecq: Pushing the limit of post-training quantization by block reconstruction.
\newblock In \emph{International Conference on Learning Representations (ICLR)}, 2021.

\bibitem[Lin et~al.(2023)Lin, Tang, Tang, Yang, Dang, and Han]{lin2023awq}
Ji~Lin, Jiaming Tang, Haotian Tang, Shang Yang, Xingyu Dang, and Song Han.
\newblock Awq: Activation-aware weight quantization for llm compression and acceleration.
\newblock \emph{arXiv preprint arXiv:2306.00978}, 2023.

\bibitem[Liu et~al.(2023{\natexlab{a}})Liu, Gong, Wei, Dong, Cai, and Zhuang]{liu2023qllm}
Jing Liu, Ruihao Gong, Xiuying Wei, Zhiwei Dong, Jianfei Cai, and Bohan Zhuang.
\newblock Qllm: Accurate and efficient low-bitwidth quantization for large language models.
\newblock \emph{arXiv preprint arXiv:2310.08041}, 2023{\natexlab{a}}.

\bibitem[Liu et~al.(2023{\natexlab{b}})Liu, Liu, Huang, Dong, and Cheng]{liu2023llmfp4}
Shih-yang Liu, Zechun Liu, Xijie Huang, Pingcheng Dong, and Kwang-Ting Cheng.
\newblock Llm-fp4: 4-bit floating-point quantized transformers.
\newblock \emph{arXiv preprint arXiv:2310.16836}, 2023{\natexlab{b}}.

\bibitem[Liu et~al.(2023{\natexlab{c}})Liu, Oguz, Zhao, Chang, Stock, Mehdad, Shi, Krishnamoorthi, and Chandra]{liu2023llmqat}
Zechun Liu, Barlas Oguz, Changsheng Zhao, Ernie Chang, Pierre Stock, Yashar Mehdad, Yangyang Shi, Raghuraman Krishnamoorthi, and Vikas Chandra.
\newblock Llm-qat: Data-free quantization aware training for large language models.
\newblock \emph{arXiv preprint arXiv:2305.17888}, 2023{\natexlab{c}}.

\bibitem[Liu et~al.(2024)Liu, Zhao, Iandola, Lai, Tian, Fedorov, Xiong, Chang, Shi, Krishnamoorthi, et~al.]{liu2024mobilellm}
Zechun Liu, Changsheng Zhao, Forrest Iandola, Chen Lai, Yuandong Tian, Igor Fedorov, Yunyang Xiong, Ernie Chang, Yangyang Shi, Raghuraman Krishnamoorthi, et~al.
\newblock Mobilellm: Optimizing sub-billion parameter language models for on-device use cases.
\newblock \emph{arXiv preprint arXiv:2402.14905}, 2024.

\bibitem[Merity et~al.(2016)Merity, Xiong, Bradbury, and Socher]{merity2016wiki2}
Stephen Merity, Caiming Xiong, James Bradbury, and Richard Socher.
\newblock Pointer sentinel mixture models.
\newblock \emph{arXiv preprint arXiv:1609.07843}, 2016.

\bibitem[Mihaylov et~al.(2018)Mihaylov, Clark, Khot, and Sabharwal]{mihaylov2018obqa}
Todor Mihaylov, Peter Clark, Tushar Khot, and Ashish Sabharwal.
\newblock Can a suit of armor conduct electricity? a new dataset for open book question answering.
\newblock \emph{arXiv preprint arXiv:1809.02789}, 2018.

\bibitem[Nagel et~al.(2019)Nagel, Baalen, Blankevoort, and Welling]{nagel2019data}
Markus Nagel, Mart~van Baalen, Tijmen Blankevoort, and Max Welling.
\newblock Data-free quantization through weight equalization and bias correction.
\newblock In \emph{Proceedings of the IEEE/CVF International Conference on Computer Vision}, pages 1325--1334, 2019.

\bibitem[Nagel et~al.(2020)Nagel, Amjad, Van~Baalen, Louizos, and Blankevoort]{adaround}
Markus Nagel, Rana~Ali Amjad, Mart Van~Baalen, Christos Louizos, and Tijmen Blankevoort.
\newblock Up or down? {A}daptive rounding for post-training quantization.
\newblock In \emph{International Conference on Machine Learning (ICML)}, 2020.

\bibitem[Nishimori and Akaho(2005)]{nishimori2005learning}
Yasunori Nishimori and Shotaro Akaho.
\newblock Learning algorithms utilizing quasi-geodesic flows on the stiefel manifold.
\newblock \emph{Neurocomputing}, 67:\penalty0 106--135, 2005.

\bibitem[Raffel et~al.(2020)Raffel, Shazeer, Roberts, Lee, Narang, Matena, Zhou, Li, and Liu]{raffel2020c4}
Colin Raffel, Noam Shazeer, Adam Roberts, Katherine Lee, Sharan Narang, Michael Matena, Yanqi Zhou, Wei Li, and Peter~J Liu.
\newblock Exploring the limits of transfer learning with a unified text-to-text transformer.
\newblock \emph{The Journal of Machine Learning Research}, 21\penalty0 (1):\penalty0 5485--5551, 2020.

\bibitem[Roziere et~al.(2023)Roziere, Gehring, Gloeckle, Sootla, Gat, Tan, Adi, Liu, Remez, Rapin, et~al.]{roziere2023code}
Baptiste Roziere, Jonas Gehring, Fabian Gloeckle, Sten Sootla, Itai Gat, Xiaoqing~Ellen Tan, Yossi Adi, Jingyu Liu, Tal Remez, J{\'e}r{\'e}my Rapin, et~al.
\newblock Code llama: Open foundation models for code.
\newblock \emph{arXiv preprint arXiv:2308.12950}, 2023.

\bibitem[Sakaguchi et~al.(2021)Sakaguchi, Bras, Bhagavatula, and Choi]{sakaguchi2021winogrande}
Keisuke Sakaguchi, Ronan~Le Bras, Chandra Bhagavatula, and Yejin Choi.
\newblock Winogrande: An adversarial winograd schema challenge at scale.
\newblock \emph{Communications of the ACM}, 64\penalty0 (9):\penalty0 99--106, 2021.

\bibitem[Sap et~al.(2019)Sap, Rashkin, Chen, LeBras, and Choi]{sap2019siqa}
Maarten Sap, Hannah Rashkin, Derek Chen, Ronan LeBras, and Yejin Choi.
\newblock Socialiqa: Commonsense reasoning about social interactions.
\newblock \emph{arXiv preprint arXiv:1904.09728}, 2019.

\bibitem[Shao et~al.(2023)Shao, Chen, Zhang, Xu, Zhao, Li, Zhang, Gao, Qiao, and Luo]{shao2023omniquant}
Wenqi Shao, Mengzhao Chen, Zhaoyang Zhang, Peng Xu, Lirui Zhao, Zhiqian Li, Kaipeng Zhang, Peng Gao, Yu~Qiao, and Ping Luo.
\newblock Omniquant: Omnidirectionally calibrated quantization for large language models.
\newblock \emph{arXiv preprint arXiv:2308.13137}, 2023.

\bibitem[Team et~al.(2023)Team, Anil, Borgeaud, Wu, Alayrac, Yu, Soricut, Schalkwyk, Dai, Hauth, et~al.]{team2023gemini}
Gemini Team, Rohan Anil, Sebastian Borgeaud, Yonghui Wu, Jean-Baptiste Alayrac, Jiahui Yu, Radu Soricut, Johan Schalkwyk, Andrew~M Dai, Anja Hauth, et~al.
\newblock Gemini: a family of highly capable multimodal models.
\newblock \emph{arXiv preprint arXiv:2312.11805}, 2023.

\bibitem[Thirunavukarasu et~al.(2023)Thirunavukarasu, Ting, Elangovan, Gutierrez, Tan, and Ting]{thirunavukarasu2023large}
Arun~James Thirunavukarasu, Darren Shu~Jeng Ting, Kabilan Elangovan, Laura Gutierrez, Ting~Fang Tan, and Daniel Shu~Wei Ting.
\newblock Large language models in medicine.
\newblock \emph{Nature medicine}, 29\penalty0 (8):\penalty0 1930--1940, 2023.

\bibitem[Touvron et~al.(2023{\natexlab{a}})Touvron, Lavril, Izacard, Martinet, Lachaux, Lacroix, Rozi{\`e}re, Goyal, Hambro, Azhar, et~al.]{llama}
Hugo Touvron, Thibaut Lavril, Gautier Izacard, Xavier Martinet, Marie-Anne Lachaux, Timoth{\'e}e Lacroix, Baptiste Rozi{\`e}re, Naman Goyal, Eric Hambro, Faisal Azhar, et~al.
\newblock Llama: Open and efficient foundation language models.
\newblock \emph{arXiv preprint arXiv:2302.13971}, 2023{\natexlab{a}}.

\bibitem[Touvron et~al.(2023{\natexlab{b}})Touvron, Martin, Stone, Albert, Almahairi, Babaei, Bashlykov, Batra, Bhargava, Bhosale, et~al.]{touvron2023llama}
Hugo Touvron, Louis Martin, Kevin Stone, Peter Albert, Amjad Almahairi, Yasmine Babaei, Nikolay Bashlykov, Soumya Batra, Prajjwal Bhargava, Shruti Bhosale, et~al.
\newblock Llama 2: Open foundation and fine-tuned chat models.
\newblock \emph{arXiv preprint arXiv:2307.09288}, 2023{\natexlab{b}}.

\bibitem[Tseng et~al.(2024)Tseng, Chee, Sun, Kuleshov, and De~Sa]{tseng2024quip}
Albert Tseng, Jerry Chee, Qingyao Sun, Volodymyr Kuleshov, and Christopher De~Sa.
\newblock Quip\#: Even better llm quantization with hadamard incoherence and lattice codebooks.
\newblock \emph{arXiv preprint arXiv:2402.04396}, 2024.

\bibitem[van Baalen et~al.(2023)van Baalen, Andrey~Kuzmin, Couperus, Bastoul, Mahurin, Blankevoort, and Whatmough]{baalen2023gptvq}
Mart van Baalen, Markus~Nagel Andrey~Kuzmin, Peter Couperus, Cedric Bastoul, Eric Mahurin, Tijmen Blankevoort, and Paul Whatmough.
\newblock Gptvq: The blessing of dimensionality for llm quantization.
\newblock 2023.

\bibitem[Wei et~al.(2022)Wei, Zhang, Zhang, Gong, Zhang, Zhang, Yu, and Liu]{wei2022outlier}
Xiuying Wei, Yunchen Zhang, Xiangguo Zhang, Ruihao Gong, Shanghang Zhang, Qi~Zhang, Fengwei Yu, and Xianglong Liu.
\newblock Outlier suppression: Pushing the limit of low-bit transformer language models.
\newblock \emph{arXiv preprint arXiv:2209.13325}, 2022.

\bibitem[Wei et~al.(2023)Wei, Zhang, Li, Zhang, Gong, Guo, and Liu]{wei2023outlier}
Xiuying Wei, Yunchen Zhang, Yuhang Li, Xiangguo Zhang, Ruihao Gong, Jinyang Guo, and Xianglong Liu.
\newblock Outlier suppression+: Accurate quantization of large language models by equivalent and optimal shifting and scaling.
\newblock \emph{arXiv preprint arXiv:2304.09145}, 2023.

\bibitem[Xiao et~al.(2022)Xiao, Lin, Seznec, Wu, Demouth, and Han]{xiao2022smoothquant}
Guangxuan Xiao, Ji~Lin, Mickael Seznec, Hao Wu, Julien Demouth, and Song Han.
\newblock Smoothquant: Accurate and efficient post-training quantization for large language models.
\newblock In \emph{CVPR}, 2022.

\bibitem[Xiao et~al.(2023)Xiao, Lin, Seznec, Wu, Demouth, and Han]{xiao2023smoothquant}
Guangxuan Xiao, Ji~Lin, Mickael Seznec, Hao Wu, Julien Demouth, and Song Han.
\newblock Smoothquant: Accurate and efficient post-training quantization for large language models.
\newblock In \emph{International Conference on Machine Learning}, pages 38087--38099. PMLR, 2023.

\bibitem[Yelysei~Bondarenko(2023)]{bondarenko2023outliers}
Tijmen~Blankevoort Yelysei~Bondarenko, Markus~Nagel.
\newblock Quantizable transformers: Removing outliers by helping attention heads do nothing.
\newblock 2023.

\bibitem[Zellers et~al.(2019)Zellers, Holtzman, Bisk, Farhadi, and Choi]{zellers2019hellaswag}
Rowan Zellers, Ari Holtzman, Yonatan Bisk, Ali Farhadi, and Yejin Choi.
\newblock Hellaswag: Can a machine really finish your sentence?
\newblock \emph{arXiv preprint arXiv:1905.07830}, 2019.

\bibitem[Zhao et~al.(2023)Zhao, Lin, Zhu, Ye, Chen, Zheng, Ceze, Krishnamurthy, Chen, and Kasikci]{zhao2023atom}
Yilong Zhao, Chien-Yu Lin, Kan Zhu, Zihao Ye, Lequn Chen, Size Zheng, Luis Ceze, Arvind Krishnamurthy, Tianqi Chen, and Baris Kasikci.
\newblock Atom: Low-bit quantization for efficient and accurate llm serving.
\newblock \emph{arXiv preprint arXiv:2310.19102}, 2023.

\end{thebibliography}
    \bibliographystyle{plainnat}
    }

\newpage
\appendix

\section{Appendix / supplemental material}

\subsection{Complete results of main result table}
In Tables~\ref{tab:appendix_main_llama2}, ~\ref{tab:appendix_main_llama3} and ~\ref{tab:appendix_mistral}, we show the complete results of Table~\ref{tab:main}. We compare the accuracy on eight zero-shot commonsense reasoning tasks including ARC-easy, ARC-challenge~\citep{clark2018arc}, BoolQ~\citep{clark2019boolq}, PIQA~\citep{bisk2020piqa}, SIQA~\citep{sap2019siqa}, HellaSwag~\citep{zellers2019hellaswag}, OBQA~\citep{mihaylov2018obqa}, and WinoGrande~\citep{sakaguchi2021winogrande} as well as the perplexity score on WikiText2 testset~\citep{merity2016wiki2}. We compare our results with previous works including SmoothQuant\citep{xiao2022smoothquant}, LLM-QAT\citep{liu2023llmqat}, GPTQ~\citep{frantar2022gptq}, OmniQuant~\citep{shao2023omniquant}, QuIP\#~\citep{tseng2024quip}.
\begin{table*}[t!]
\renewcommand\arraystretch{0.8}
\centering
\caption{\small Complete comparison of the perplexity score on WikiText2 and averaged accuracy on Zero-shot Common Sense Reasoning tasks on \textbf{LLaMA-2}.}
\vspace{-4.8em}
\label{tab:appendix_main_llama2}
\setlength{\tabcolsep}{1mm}
{\resizebox{0.9\textwidth}{!}{
\begin{tabular}{c|c|l|ccccccccc|c}
& & & & & & & & & & &\\
& & & & & & & & & & &\\
& & & & & & & & & & &\\
& & & & & & & & & & &\\
& & & & & & & & & & &\\
& & & & & & & & & & &\\
\noalign{\vspace{0.2em}}\hline\noalign{\vspace{0.1em}}
\noalign{\vspace{0.1em}}\hline\noalign{\vspace{0.2em}}
\multirow{2}{*}{\textbf{Model}} & \textbf{\#Bits} & \multirow{2}{*}{\textbf{Method}} & \textbf{ARC-e} & \textbf{ARC-c} & \textbf{BoolQ} & \textbf{PIQA} & \textbf{SIQA} & \textbf{HellaS.} & \textbf{OBQA} & \textbf{WinoG.} & \textbf{Avg.}  & \textbf{Wiki2} \\ 
& W-A-KV & & ($\uparrow$) & ($\uparrow$) & ($\uparrow$) & ($\uparrow$) & ($\uparrow$) & ($\uparrow$) & ($\uparrow$) & ($\uparrow$) & ($\uparrow$) & ($\downarrow$) \\
\noalign{\vspace{0.2em}}\hline\noalign{\vspace{0.2em}} 
\multirow{30}{*}{7B} & 16-16-16 & Full Precision & 75.0 & 50.8 & 77.3 & 78.9 & 48.5 & 76.0 & 59.3 & 69.5 & 66.9 & 5.5 \\
\noalign{\vspace{0.2em}}\cdashline{2-13}\noalign{\vspace{0.2em}}
& \multirow{8}{*}{4-8-16} & RTN & 70.9 & 44.3 & 73.5 & 76.8 & 46.0 & 70.3 & 51.8 & 65.9 & 62.4 & 7.9 \\
&  & SmoothQuant & 65.8 & 41.7 & 67.3 & 75.6 & 44.5 & 67.1 & 45.8 & 63.5 & 58.9 & 7.5 \\
&  & LLM-QAT & 73.6 & 49.0 & 72.4 & 78.2 & 47.8 & 74.0 & 56.1 & 67.7 & 64.8 & 11.4 \\
&  & AWQ (w4) & -- & -- & -- & -- & -- & -- & -- & -- & -- & 6.2 \\
&  & OmniQuant (w4) & -- & -- & -- & -- & -- & -- & -- & -- & -- & 5.7 \\
&  & QuIP\# (w4) & -- & -- & -- & -- & -- & -- & -- & -- & -- & 5.6 \\
&  & GPTQ & 73.7 & 47.5 & 74.8 & 77.7 & 46.4 & 74.1 & 55.7 & 69.3 & 64.9 & 20.2 \\
\rowcolor{gray!20} \cellcolor{white} & \cellcolor{white} & $\ourse{}$ & 73.6 & 49.4 & 76.0 & 79.0 & 47.8 & 75.0 & 56.1 & 68.8 & 65.7 & 5.8 \\
\rowcolor{gray!20} \cellcolor{white} & \cellcolor{white} & $\oursh{}$ & 74.0 & 50.1 & 74.4 & 78.9 & 47.6 & 74.8 & 56.7 & 68.9 & 65.7 & 5.7 \\
\noalign{\vspace{0.1em}}\cdashline{2-13}\noalign{\vspace{0.1em}}
& \multirow{6}{*}{4-8-8} & RTN & 71.1 & 44.3 & 73.2 & 76.8 & 45.8 & 70.3 & 52.3 & 65.8 & 62.5 & 7.9 \\
&  & SmoothQuant & 65.8 & 40.8 & 66.4 & 76.3 & 43.7 & 66.9 & 46.0 & 64.5 & 58.8 & 7.5 \\
&  & LLM-QAT & 73.5 & 48.3 & 72.4 & 78.1 & 47.4 & 74.0 & 55.3 & 68.0 & 64.6 & 11.4 \\
&  & GPTQ & 73.7 & 48.0 & 74.2 & 78.1 & 46.6 & 73.9 & 55.1 & 68.5 & 64.8 & 20.2 \\
\rowcolor{gray!20} \cellcolor{white} & \cellcolor{white} & $\ourse{}$ & 75.1 & 49.8 & 74.7 & 78.2 & 47.8 & 75.0 & 57.6 & 67.7 & 65.8 & 5.8 \\
\rowcolor{gray!20} \cellcolor{white} & \cellcolor{white} & $\oursh{}$ & 73.4 & 49.6 & 76.0 & 78.4 & 47.7 & 74.6 & 56.2 & 70.3 & 65.8 & 5.7 \\
\noalign{\vspace{0.1em}}\cdashline{2-13}\noalign{\vspace{0.1em}}
& \multirow{6}{*}{4-4-16} & RTN & 26.6 & 22.1 & 44.3 & 50.9 & 38.9 & 26.2 & 26.6 & 49.4 & 35.6 & 2,167.2 \\
&  & SmoothQuant & 37.8 & 27.1 & 51.9 & 59.4 & 40.2 & 34.3 & 31.6 & 52.4 & 41.8 & 254.5 \\
&  & LLM-QAT & 46.2 & 32.4 & 61.8 & 62.0 & 41.3 & 47.6 & 36.1 & 54.7 & 47.8 & 12.9 \\
&  & GPTQ & 27.6 & 24.9 & 47.4 & 50.7 & 38.6 & 26.9 & 28.3 & 49.9 & 36.8 & 8,949.0 \\
\rowcolor{gray!20} \cellcolor{white} & \cellcolor{white} & $\ourse{}$ & 61.0 & 39.4 & 66.0 & 72.6 & 44.5 & 66.1 & 45.1 & 61.6 & 57.0 & 9.2 \\
\rowcolor{gray!20} \cellcolor{white} & \cellcolor{white} & $\oursh{}$ & 72.1 & 47.5 & 74.4 & 77.0 & 47.3 & 73.2 & 54.4 & 66.9 & 64.1 & 5.9 \\
\noalign{\vspace{0.1em}}\cdashline{2-13}\noalign{\vspace{0.1em}}
& \multirow{6}{*}{4-4-4} & RTN & 27.1 & 24.4 & 44.8 & 51.4 & 39.4 & 26.7 & 33.0 & 50.0 & 37.1 & 2,382.5 \\
&  & SmoothQuant & 31.4 & 24.8 & 51.4 & 54.1 & 39.4 & 29.1 & 31.9 & 50.0 & 39.0 & 698.7 \\
&  & LLM-QAT & 42.0 & 27.7 & 59.5 & 58.9 & 41.0 & 43.1 & 33.5 & 53.3 & 44.9 & 14.9 \\
&  & GPTQ & 27.6 & 23.6 & 47.8 & 51.0 & 38.7 & 27.0 & 28.5 & 50.3 & 36.8 & 9,253.1 \\
\rowcolor{gray!20} \cellcolor{white} & \cellcolor{white} & $\ourse{}$ & 61.8 & 39.1 & 64.8 & 71.6 & 44.5 & 65.0 & 41.4 & 60.0 & 56.0 & 9.2 \\
\rowcolor{gray!20} \cellcolor{white} & \cellcolor{white} & $\oursh{}$ & 72.6 & 47.5 & 73.9 & 77.0 & 47.2 & 73.0 & 54.1 & 66.9 & 64.0 & 5.9 \\
\noalign{\vspace{0.2em}}\hline\noalign{\vspace{0.2em}}
\multirow{30}{*}{13B} & 16-16-16 & Full Precision & 75.3 & 51.4 & 79.8 & 80.4 & 50.5 & 79.8 & 56.8 & 72.5 & 68.3 & 5.0 \\
\noalign{\vspace{0.2em}}\cdashline{2-13}\noalign{\vspace{0.2em}}
& \multirow{8}{*}{4-8-16} & RTN & 63.1 & 39.9 & 68.7 & 74.0 & 46.2 & 59.7 & 45.5 & 61.5 & 57.3 & 6.7 \\
&  & SmoothQuant & 71.7 & 46.3 & 72.0 & 78.2 & 47.3 & 72.8 & 51.2 & 69.2 & 63.6 & 6.1 \\
&  & LLM-QAT & 75.3 & 49.7 & 79.0 & 80.0 & 50.3 & 77.4 & 56.3 & 71.6 & 67.5 & 14.5 \\
&  & AWQ (w4) & -- & -- & -- & -- & -- & -- & -- & -- & -- & 5.1 \\
&  & OmniQuant (w4) & -- & -- & -- & -- & -- & -- & -- & -- & -- & 5.0 \\
&  & QuIP\# (w4) & -- & -- & -- & -- & -- & -- & -- & -- & -- & 5.0 \\
&  & GPTQ & 74.2 & 49.2 & 75.3 & 78.4 & 48.8 & 74.1 & 53.4 & 68.4 & 65.2 & 5.9 \\
\rowcolor{gray!20} \cellcolor{white} & \cellcolor{white} & $\ourse{}$ & 76.5 & 52.0 & 81.5 & 80.0 & 49.9 & 78.8 & 54.8 & 72.4 & 68.2 & 5.1 \\
\rowcolor{gray!20} \cellcolor{white} & \cellcolor{white} & $\oursh{}$ & 76.2 & 50.6 & 80.1 & 80.1 & 49.8 & 78.5 & 58.0 & 71.7 & 68.1 & 5.0 \\
\noalign{\vspace{0.1em}}\cdashline{2-13}\noalign{\vspace{0.1em}}
& \multirow{6}{*}{4-8-8} & RTN & 63.2 & 40.3 & 69.0 & 74.3 & 46.1 & 59.5 & 46.2 & 61.9 & 57.6 & 6.7 \\
&  & SmoothQuant & 73.3 & 45.3 & 71.9 & 78.8 & 47.6 & 72.7 & 49.6 & 67.7 & 63.4 & 6.1 \\
&  & LLM-QAT & 75.0 & 48.8 & 79.2 & 80.3 & 50.7 & 77.7 & 56.1 & 72.3 & 67.5 & 14.2 \\
&  & GPTQ & 74.1 & 48.8 & 75.1 & 78.1 & 48.8 & 74.1 & 53.6 & 69.5 & 65.3 & 5.9 \\
\rowcolor{gray!20} \cellcolor{white} & \cellcolor{white} & $\ourse{}$ & 76.8 & 52.1 & 80.8 & 80.5 & 49.9 & 78.6 & 55.8 & 70.6 & 68.1 & 5.1 \\
\rowcolor{gray!20} \cellcolor{white} & \cellcolor{white} & $\oursh{}$ & 76.7 & 51.2 & 80.4 & 80.5 & 49.4 & 78.6 & 57.4 & 71.5 & 68.2 & 5.1 \\
\noalign{\vspace{0.1em}}\cdashline{2-13}\noalign{\vspace{0.1em}}
& \multirow{6}{*}{4-4-16} & RTN & 26.0 & 26.0 & 40.6 & 49.7 & 38.7 & 26.0 & 25.4 & 49.9 & 35.3 & 7,216.7 \\
&  & SmoothQuant & 45.2 & 27.1 & 55.4 & 62.5 & 40.5 & 44.3 & 33.4 & 50.8 & 44.9 & 34.5 \\
&  & LLM-QAT & 26.0 & 23.7 & 37.8 & 49.2 & 39.5 & 26.3 & 23.8 & 48.2 & 34.3 & 3,889.9 \\
&  & GPTQ & 26.6 & 24.7 & 37.9 & 49.3 & 39.2 & 26.2 & 27.7 & 50.3 & 35.2 & 5,245.3 \\
\rowcolor{gray!20} \cellcolor{white} & \cellcolor{white} & $\ourse{}$ & 68.5 & 43.0 & 72.1 & 75.4 & 48.5 & 71.2 & 51.0 & 64.6 & 61.8 & 7.2 \\
\rowcolor{gray!20} \cellcolor{white} & \cellcolor{white} & $\oursh{}$ & 75.9 & 50.8 & 78.1 & 79.5 & 49.4 & 77.5 & 55.2 & 70.8 & 67.2 & 5.2 \\
\noalign{\vspace{0.1em}}\cdashline{2-13}\noalign{\vspace{0.1em}}
& \multirow{6}{*}{4-4-4} & RTN & 26.1 & 24.3 & 40.3 & 48.7 & 39.6 & 25.8 & 29.2 & 49.6 & 35.5 & 7,428.8 \\
&  & SmoothQuant & 36.9 & 24.8 & 49.4 & 57.2 & 39.6 & 33.3 & 31.2 & 51.7 & 40.5 & 56.6 \\
&  & LLM-QAT & 26.3 & 24.6 & 37.8 & 48.8 & 39.3 & 26.3 & 26.8 & 50.4 & 35.0 & 3,777.5 \\
&  & GPTQ & 26.6 & 24.1 & 37.9 & 48.8 & 38.9 & 26.1 & 29.3 & 50.1 & 35.2 & 5,237.1 \\
\rowcolor{gray!20} \cellcolor{white} & \cellcolor{white} & $\ourse{}$ & 67.1 & 39.7 & 72.5 & 74.7 & 47.4 & 71.1 & 47.8 & 65.3 & 60.7 & 7.1 \\
\rowcolor{gray!20} \cellcolor{white} & \cellcolor{white} & $\oursh{}$ & 75.7 & 50.5 & 79.3 & 79.5 & 49.1 & 77.1 & 53.8 & 69.9 & 66.9 & 5.3 \\
\noalign{\vspace{0.2em}}\hline\noalign{\vspace{0.2em}}
\multirow{24}{*}{70B} & 16-16-16 & Full Precision & 80.2 & 60.5 & 85.1 & 82.8 & 50.8 & 84.3 & 59.0 & 80.6 & 72.9 & 3.3 \\
\noalign{\vspace{0.2em}}\cdashline{2-13}\noalign{\vspace{0.2em}}
& \multirow{7}{*}{4-8-16} & RTN & 78.2 & 54.8 & 81.5 & 80.8 & 46.9 & 76.5 & 56.5 & 73.3 & 68.6 & 5.0 \\
&  & SmoothQuant & 79.4 & 57.3 & 82.4 & 82.0 & 50.3 & 81.5 & 56.2 & 75.9 & 70.6 & 4.1 \\
&  & OmniQuant (w4) & -- & -- & -- & -- & -- & -- & -- & -- & -- & 3.5 \\
&  & QuIP\# (w4) & -- & -- & -- & -- & -- & -- & -- & -- & -- & 3.4 \\
&  & GPTQ & 80.2 & 59.5 & 82.4 & 82.6 & 50.3 & 82.1 & 58.3 & 77.9 & 71.7 & 4.3 \\
\rowcolor{gray!20} \cellcolor{white} & \cellcolor{white} & $\ourse{}$ & 80.0 & 59.2 & 84.4 & 82.6 & 50.3 & 82.8 & 59.7 & 78.1 & 72.1 & 3.7 \\
\rowcolor{gray!20} \cellcolor{white} & \cellcolor{white} & $\oursh{}$ & 80.2 & 59.9 & 85.0 & 82.5 & 50.4 & 83.9 & 60.1 & 79.3 & 72.7 & 3.5 \\
\noalign{\vspace{0.1em}}\cdashline{2-13}\noalign{\vspace{0.1em}}
& \multirow{5}{*}{4-8-8} & RTN & 78.3 & 53.9 & 81.4 & 81.4 & 47.3 & 76.7 & 56.0 & 72.6 & 68.4 & 5.0 \\
&  & SmoothQuant & 80.0 & 57.8 & 81.6 & 81.6 & 48.9 & 81.5 & 56.6 & 75.8 & 70.5 & 4.1 \\
&  & GPTQ & 79.6 & 60.3 & 82.4 & 82.2 & 49.9 & 82.2 & 58.5 & 77.3 & 71.6 & 4.3 \\
\rowcolor{gray!20} \cellcolor{white} & \cellcolor{white} & $\ourse{}$ & 80.4 & 60.3 & 84.4 & 81.8 & 49.8 & 82.8 & 59.1 & 79.0 & 72.2 & 3.7 \\
\rowcolor{gray!20} \cellcolor{white} & \cellcolor{white} & $\oursh{}$ & 80.4 & 59.7 & 85.2 & 82.6 & 50.4 & 83.8 & 59.9 & 79.8 & 72.7 & 3.5 \\
\noalign{\vspace{0.1em}}\cdashline{2-13}\noalign{\vspace{0.1em}}
& \multirow{5}{*}{4-4-16} & RTN & 26.0 & 23.2 & 43.5 & 48.9 & 37.0 & 26.0 & 25.6 & 50.5 & 35.1 & 2e5 \\
&  & SmoothQuant & 69.5 & 71.7 & 29.0 & 66.6 & 73.1 & 45.1 & 67.4 & 39.4 & 57.7 & 57.1 \\
&  & GPTQ & 25.3 & 25.8 & 45.7 & 50.1 & 36.4 & 25.8 & 24.6 & 50.0 & 35.5 & 2e6 \\
\rowcolor{gray!20} \cellcolor{white} & \cellcolor{white} & $\ourse{}$ & 66.8 & 42.4 & 72.9 & 74.0 & 46.7 & 73.2 & 48.2 & 63.9 & 61.0 & 7.3 \\
\rowcolor{gray!20} \cellcolor{white} & \cellcolor{white} & $\oursh{}$ & 78.4 & 57.0 & 82.7 & 81.4 & 50.2 & 83.0 & 58.5 & 77.0 & 71.0 & 3.8 \\
\noalign{\vspace{0.1em}}\cdashline{2-13}\noalign{\vspace{0.1em}}
& \multirow{5}{*}{4-4-4} & RTN & 25.5 & 24.5 & 43.2 & 50.2 & 36.7 & 26.6 & 24.2 & 49.3 & 35.0 & 2e5 \\
&  & SmoothQuant & 68.1 & 31.9 & 65.8 & 72.0 & 43.5 & 64.2 & 38.2 & 63.1 & 55.9 & 10.5 \\
&  & GPTQ & 26.1 & 25.2 & 45.7 & 49.5 & 36.8 & 26.0 & 25.4 & 50.2 & 35.6 & 1e6 \\
\rowcolor{gray!20} \cellcolor{white} & \cellcolor{white} & $\ourse{}$ & 68.2 & 42.0 & 74.1 & 73.8 & 46.9 & 74.3 & 50.0 & 66.8 & 62.0 & 7.4 \\
\rowcolor{gray!20} \cellcolor{white} & \cellcolor{white} & $\oursh{}$ & 78.3 & 57.6 & 82.1 & 81.7 & 50.1 & 82.9 & 59.8 & 77.3 & 71.2 & 3.8 \\
\noalign{\vspace{0.2em}}\hline\noalign{\vspace{0.2em}}
\hline
\end{tabular}}}
\vspace{2em}
\end{table*}
\begin{table*}[t!]
\renewcommand\arraystretch{0.8}
\centering
\caption{\small Complete omparison of the perplexity score on WikiText2 and averaged accuracy on Zero-shot Common Sense Reasoning tasks on \textbf{LLaMA-3}.}
\vspace{-4em}
\label{tab:appendix_main_llama3}
\setlength{\tabcolsep}{1mm}
{\resizebox{0.9\textwidth}{!}{
\begin{tabular}{c|c|l|ccccccccc|c}
& & & & & & & & & & &\\
& & & & & & & & & & &\\
& & & & & & & & & & &\\
& & & & & & & & & & &\\
& & & & & & & & & & &\\
\noalign{\vspace{0.2em}}\hline\noalign{\vspace{0.1em}}
\noalign{\vspace{0.1em}}\hline\noalign{\vspace{0.2em}}
\multirow{2}{*}{\textbf{Model}} & \textbf{\#Bits} & \multirow{2}{*}{\textbf{Method}} & \textbf{ARC-e} & \textbf{ARC-c} & \textbf{BoolQ} & \textbf{PIQA} & \textbf{SIQA} & \textbf{HellaS.} & \textbf{OBQA} & \textbf{WinoG.} & \textbf{Avg.}  & \textbf{Wiki2} \\ 
& W-A-KV & & ($\uparrow$) & ($\uparrow$) & ($\uparrow$) & ($\uparrow$) & ($\uparrow$) & ($\uparrow$) & ($\uparrow$) & ($\uparrow$) & ($\uparrow$) & ($\downarrow$) \\
\noalign{\vspace{0.2em}}\hline\noalign{\vspace{0.2em}}
 & 16-16-16 & Full Precision & 65.2 & 38.7 & 69.5 & 75.3 & 44.8 & 60.7 & 40.2 & 60.9 & 56.9 & 13.4 \\
\noalign{\vspace{0.1em}}\cdashline{2-13}\noalign{\vspace{0.1em}}
 & \multirow{6}{*}{4-8-16} & RTN & 62.4 & 39.7 & 66.3 & 72.1 & 44.6 & 56.6 & 42.8 & 58.6 & 55.4 & 20.7 \\
 &  & SmoothQuant & 47.6 & 30.7 & 59.6 & 64.9 & 41.7 & 47.6 & 31.5 & 52.9 & 47.1 & 108.2 \\
 &  & LLM-QAT & 59.6 & 37.8 & 61.7 & 72.5 & 43.1 & 57.2 & 37.1 & 56.2 & 53.2 & 21.0 \\
 &  & GPTQ & 61.7 & 38.5 & 65.6 & 71.4 & 43.9 & 56.4 & 44.1 & 58.7 & 55.0 & 17.3 \\
 \rowcolor{gray!20} \cellcolor{white} & \cellcolor{white} & $\ourse{}$ & 60.9 & 39.5 & 65.9 & 73.2 & 46.1 & 57.7 & 44.3 & 60.3 & 56.0 & 15.3 \\
 \rowcolor{gray!20} \cellcolor{white} & \cellcolor{white} & $\oursh{}$ & 60.8 & 39.8 & 66.5 & 73.9 & 44.7 & 59.0 & 46.9 & 60.8 & 56.5 & 14.4 \\
\noalign{\vspace{0.1em}}\cdashline{2-13}\noalign{\vspace{0.1em}}
 & \multirow{6}{*}{4-8-8} & RTN & 62.6 & 40.0 & 66.7 & 72.2 & 44.4 & 56.6 & 43.0 & 59.9 & 55.7 & 20.7 \\
 &  & SmoothQuant & 48.2 & 31.5 & 59.1 & 65.4 & 41.7 & 47.2 & 31.5 & 52.0 & 47.1 & 108.6 \\
 &  & LLM-QAT & 60.0 & 36.7 & 62.2 & 73.1 & 43.0 & 57.0 & 37.7 & 55.2 & 53.1 & 21.0 \\
 &  & GPTQ & 61.7 & 38.0 & 65.4 & 71.4 & 43.5 & 56.1 & 45.3 & 57.0 & 54.8 & 17.3 \\
 \rowcolor{gray!20} \cellcolor{white} & \cellcolor{white} & $\ourse{}$ & 61.2 & 40.6 & 64.9 & 72.7 & 45.1 & 58.3 & 43.2 & 59.9 & 55.7 & 15.3 \\
 \rowcolor{gray!20} \cellcolor{white} 1B & \cellcolor{white} & $\oursh{}$ & 59.2 & 37.3 & 66.4 & 73.6 & 44.9 & 59.1 & 46.3 & 59.2 & 55.8 & 14.3 \\
\noalign{\vspace{0.1em}}\cdashline{2-13}\noalign{\vspace{0.1em}}
 & \multirow{6}{*}{4-4-16} & RTN & 37.8 & 28.3 & 51.2 & 56.4 & 40.0 & 35.9 & 28.9 & 51.4 & 41.2 & 137.5 \\
 &  & SmoothQuant & 32.3 & 26.4 & 46.3 & 54.7 & 39.7 & 28.7 & 27.0 & 48.0 & 37.9 & 2,027.8 \\
 &  & LLM-QAT & 39.3 & 28.5 & 55.6 & 58.9 & 40.9 & 32.7 & 28.1 & 52.0 & 42.0 & 62.1 \\
 &  & GPTQ & 36.8 & 27.1 & 56.0 & 56.6 & 41.2 & 36.0 & 27.9 & 51.5 & 41.6 & 107.6 \\
 \rowcolor{gray!20} \cellcolor{white} & \cellcolor{white} & $\ourse{}$ & 44.8 & 29.7 & 61.2 & 59.7 & 40.2 & 41.0 & 32.4 & 49.8 & 44.8 & 48.4 \\
 \rowcolor{gray!20} \cellcolor{white} & \cellcolor{white} & $\oursh{}$ & 59.3 & 37.1 & 64.6 & 69.9 & 44.4 & 55.4 & 41.2 & 56.0 & 53.5 & 15.3 \\
\noalign{\vspace{0.1em}}\cdashline{2-13}\noalign{\vspace{0.1em}}
 & \multirow{6}{*}{4-4-4} & RTN & 37.6 & 27.6 & 49.3 & 56.4 & 40.7 & 35.1 & 27.0 & 51.5 & 40.6 & 160.4 \\
 &  & SmoothQuant & 30.0 & 26.3 & 41.8 & 51.6 & 39.0 & 26.9 & 26.8 & 49.5 & 36.5 & 2,599.6 \\
 &  & LLM-QAT & 37.7 & 26.7 & 55.7 & 57.9 & 40.6 & 32.0 & 31.3 & 50.5 & 41.5 & 76.2 \\
 &  & GPTQ & 37.8 & 29.0 & 53.9 & 56.8 & 39.9 & 34.7 & 29.7 & 51.3 & 41.6 & 124.6 \\
 \rowcolor{gray!20} \cellcolor{white} & \cellcolor{white} & $\ourse{}$ & 45.4 & 30.7 & 59.2 & 60.7 & 41.4 & 40.8 & 32.6 & 51.4 & 45.3 & 47.7 \\
 \rowcolor{gray!20} \cellcolor{white} & \cellcolor{white} & $\oursh{}$ & 59.4 & 39.4 & 64.4 & 68.9 & 43.4 & 54.6 & 41.4 & 55.9 & 53.4 & 15.9 \\
\noalign{\vspace{0.2em}}\hline\noalign{\vspace{0.2em}}
 & 16-16-16 & Full Precision & 68.9 & 47.6 & 79.0 & 76.0 & 52.1 & 71.0 & 50.2 & 66.6 & 63.9 & 10.7 \\
\noalign{\vspace{0.1em}}\cdashline{2-13}\noalign{\vspace{0.1em}}
 & \multirow{6}{*}{4-8-16} & RTN & 60.2 & 42.6 & 70.9 & 72.6 & 49.7 & 66.2 & 43.6 & 62.7 & 58.6 & 29.0 \\
 &  & SmoothQuant & 59.8 & 40.7 & 59.2 & 73.8 & 46.9 & 65.5 & 40.7 & 58.5 & 55.6 & 288.5 \\
 &  & LLM-QAT & 64.7 & 46.1 & 74.1 & 75.4 & 49.3 & 69.9 & 45.3 & 61.4 & 60.8 & 41.1 \\
 &  & GPTQ & 60.8 & 41.4 & 71.9 & 73.6 & 47.7 & 65.9 & 43.4 & 65.0 & 58.7 & 25.2 \\
 \rowcolor{gray!20} \cellcolor{white} & \cellcolor{white} & $\ourse{}$ & 65.9 & 44.2 & 74.9 & 74.8 & 48.2 & 68.3 & 48.8 & 65.9 & 61.4 & 11.6 \\
 \rowcolor{gray!20} \cellcolor{white} & \cellcolor{white} & $\oursh{}$ & 66.8 & 47.2 & 78.4 & 76.0 & 50.8 & 69.2 & 50.2 & 66.7 & 63.2 & 11.5 \\
\noalign{\vspace{0.1em}}\cdashline{2-13}\noalign{\vspace{0.1em}}
 & \multirow{6}{*}{4-8-8} & RTN & 60.2 & 41.3 & 71.3 & 73.1 & 49.6 & 66.2 & 42.6 & 63.0 & 58.4 & 28.8 \\
 &  & SmoothQuant & 59.5 & 39.3 & 57.9 & 73.5 & 46.6 & 65.3 & 41.9 & 60.1 & 55.5 & 281.3 \\
 &  & LLM-QAT & 65.2 & 45.1 & 74.5 & 76.1 & 49.1 & 69.6 & 43.9 & 60.7 & 60.5 & 39.3 \\
 &  & GPTQ & 61.0 & 42.0 & 72.5 & 72.7 & 47.9 & 66.3 & 43.4 & 63.6 & 58.7 & 24.1 \\
 \rowcolor{gray!20} \cellcolor{white} & \cellcolor{white} & $\ourse{}$ & 65.2 & 45.7 & 76.1 & 75.8 & 48.7 & 69.4 & 47.9 & 65.5 & 61.8 & 11.7 \\
 \rowcolor{gray!20} \cellcolor{white} 3B & \cellcolor{white} & $\oursh{}$ & 67.2 & 46.4 & 78.4 & 76.5 & 51.0 & 69.5 & 50.6 & 66.0 & 63.2 & 11.2 \\
\noalign{\vspace{0.1em}}\cdashline{2-13}\noalign{\vspace{0.1em}}
 & \multirow{6}{*}{4-4-16} & RTN & 41.0 & 29.8 & 43.8 & 57.3 & 41.8 & 41.4 & 31.1 & 50.9 & 42.1 & 741.9 \\
 &  & SmoothQuant & 43.6 & 30.5 & 52.8 & 58.0 & 40.4 & 37.7 & 33.1 & 52.9 & 43.6 & 372.3 \\
 &  & LLM-QAT & 47.3 & 30.9 & 60.8 & 63.8 & 42.4 & 43.2 & 35.9 & 51.1 & 46.9 & 37.6 \\
 &  & GPTQ & 42.0 & 30.0 & 44.8 & 60.1 & 41.2 & 44.7 & 34.0 & 50.5 & 43.4 & 264.4 \\
 \rowcolor{gray!20} \cellcolor{white} & \cellcolor{white} & $\ourse{}$ & 54.6 & 37.7 & 65.7 & 66.7 & 43.3 & 56.3 & 41.8 & 56.9 & 52.9 & 22.4 \\
 \rowcolor{gray!20} \cellcolor{white} & \cellcolor{white} & $\oursh{}$ & 66.3 & 43.9 & 74.2 & 75.0 & 48.9 & 67.2 & 47.1 & 65.5 & 61.0 & 11.1 \\
\noalign{\vspace{0.1em}}\cdashline{2-13}\noalign{\vspace{0.1em}}
 & \multirow{6}{*}{4-4-4} & RTN & 38.4 & 26.9 & 41.3 & 58.3 & 39.9 & 40.0 & 32.2 & 52.9 & 41.2 & 799.7 \\
 &  & SmoothQuant & 36.4 & 26.2 & 50.4 & 55.8 & 39.0 & 30.3 & 30.2 & 52.2 & 40.0 & 553.2 \\
 &  & LLM-QAT & 44.4 & 29.7 & 61.5 & 62.0 & 42.3 & 41.2 & 33.8 & 52.4 & 45.9 & 42.0 \\
 &  & GPTQ & 38.2 & 25.1 & 42.0 & 56.6 & 41.5 & 44.1 & 31.1 & 50.5 & 41.1 & 352.6 \\
 \rowcolor{gray!20} \cellcolor{white} & \cellcolor{white} & $\ourse{}$ & 58.0 & 36.0 & 67.2 & 66.9 & 43.3 & 56.8 & 40.4 & 54.5 & 52.9 & 22.4 \\
 \rowcolor{gray!20} \cellcolor{white} & \cellcolor{white} & $\oursh{}$ & 66.0 & 43.2 & 76.4 & 74.6 & 47.0 & 67.7 & 45.1 & 64.2 & 60.5 & 11.4 \\
\noalign{\vspace{0.2em}}\hline\noalign{\vspace{0.2em}}
 & 16-16-16 & Full Precision & 77.6 & 57.7 & 83.3 & 80.7 & 48.7 & 79.6 & 55.8 & 73.7 & 69.6 & 6.1 \\
\noalign{\vspace{0.1em}}\cdashline{2-13}\noalign{\vspace{0.1em}}
 & \multirow{6}{*}{4-8-16} & RTN & 73.2 & 48.1 & 76.3 & 77.1 & 46.6 & 75.5 & 54.3 & 72.5 & 65.5 & 8.2 \\
 &  & SmoothQuant & 67.5 & 41.0 & 71.9 & 74.9 & 46.6 & 70.8 & 45.8 & 69.1 & 61.0 & 10.7 \\
 &  & LLM-QAT & 77.6 & 50.6 & 81.2 & 79.0 & 47.5 & 76.0 & 53.5 & 72.4 & 67.2 & 7.7 \\
 &  & GPTQ & 71.5 & 46.8 & 76.1 & 76.6 & 47.9 & 73.9 & 52.1 & 70.7 & 64.5 & 7.2 \\
 \rowcolor{gray!20} \cellcolor{white} & \cellcolor{white} & $\ourse{}$ & 77.8 & 55.4 & 80.6 & 79.9 & 48.9 & 77.5 & 55.5 & 73.3 & 68.6 & 6.7 \\
 \rowcolor{gray!20} \cellcolor{white} & \cellcolor{white} & $\oursh{}$ & 76.5 & 54.0 & 81.5 & 79.6 & 48.6 & 78.1 & 56.4 & 72.4 & 68.4 & 6.5 \\
\noalign{\vspace{0.1em}}\cdashline{2-13}\noalign{\vspace{0.1em}}
 & \multirow{6}{*}{4-8-8} & RTN & 73.7 & 49.1 & 76.5 & 77.1 & 46.7 & 75.5 & 50.8 & 73.4 & 65.3 & 8.2 \\
 &  & SmoothQuant & 66.6 & 41.8 & 73.2 & 74.1 & 45.9 & 71.1 & 48.2 & 66.5 & 60.9 & 10.7 \\
 &  & LLM-QAT & 77.2 & 50.6 & 81.5 & 79.3 & 47.7 & 76.3 & 52.0 & 70.6 & 66.9 & 7.6 \\
 &  & GPTQ & 71.5 & 46.9 & 76.6 & 76.2 & 48.5 & 73.7 & 52.1 & 71.0 & 64.6 & 7.2 \\
 \rowcolor{gray!20} \cellcolor{white} & \cellcolor{white} & $\ourse{}$ & 77.2 & 56.2 & 81.5 & 79.2 & 48.8 & 77.2 & 56.1 & 72.9 & 68.6 & 6.7 \\
 \rowcolor{gray!20} \cellcolor{white} 8B & \cellcolor{white} & $\oursh{}$ & 77.6 & 57.4 & 81.3 & 80.2 & 48.6 & 78.1 & 55.5 & 72.0 & 68.8 & 6.5 \\
\noalign{\vspace{0.1em}}\cdashline{2-13}\noalign{\vspace{0.1em}}
 & \multirow{6}{*}{4-4-16} & RTN & 42.7 & 29.5 & 54.0 & 57.8 & 39.9 & 41.2 & 36.9 & 49.4 & 43.9 & 241.6 \\
 &  & SmoothQuant & 36.3 & 26.3 & 50.6 & 54.1 & 40.3 & 31.4 & 30.6 & 52.9 & 40.3 & 867.5 \\
 &  & LLM-QAT & 44.1 & 29.7 & 58 & 61.5 & 42.1 & 39.9 & 33 & 51.3 & 44.9 & 42.9 \\
 &  & GPTQ & 39.7 & 27.6 & 40.8 & 58.5 & 41.7 & 31.9 & 32.0 & 53.1 & 40.6 & 187.9 \\
 \rowcolor{gray!20} \cellcolor{white} & \cellcolor{white} & $\ourse{}$ & 56.5 & 35.3 & 53.3 & 68.0 & 44.5 & 59.9 & 37.5 & 59.7 & 51.9 & 18.6 \\
 \rowcolor{gray!20} \cellcolor{white} & \cellcolor{white} & $\oursh{}$ & 75 & 50.9 & 78.9 & 77.5 & 47.2 & 75.9 & 52.9 & 68.5 & 65.8 & 7.1 \\
\noalign{\vspace{0.1em}}\cdashline{2-13}\noalign{\vspace{0.1em}}
 & \multirow{6}{*}{4-4-4} & RTN & 39.5 & 27.5 & 54.6 & 57.7 & 41.4 & 39.4 & 32.6 & 51.9 & 43.1 & 260.9 \\
 &  & SmoothQuant & 33.5 & 25.1 & 49.6 & 53.1 & 40.3 & 28.8 & 29.6 & 49.6 & 38.7 & 1,530.50 \\
 &  & LLM-QAT & 40.5 & 26.6 & 52.7 & 59.9 & 42.3 & 37.5 & 33.6 & 52.7 & 43.2 & 52.5 \\
 &  & GPTQ & 40.6 & 26.5 & 40.9 & 58.0 & 41.5 & 31.9 & 33.0 & 51.8 & 40.5 & 195.8 \\
 \rowcolor{gray!20} \cellcolor{white} & \cellcolor{white} & $\ourse{}$ & 58.4 & 37.1 & 54.7 & 67.7 & 43.4 & 60.1 & 41.2 & 57.9 & 52.6 & 18.6 \\
 \rowcolor{gray!20} \cellcolor{white} & \cellcolor{white} & $\oursh{}$ & 75.1 & 51.2 & 77.2 & 77.3 & 47.6 & 75.2 & 54.1 & 66.2 & 65.5 & 7.3 \\
\noalign{\vspace{0.2em}}\hline\noalign{\vspace{0.2em}}
\hline
\end{tabular}}}
\end{table*}
\begin{table*}[t!]
\renewcommand\arraystretch{0.8}
\centering
\caption{\small Complete comparison of the perplexity score on WikiText2 and averaged accuracy on Zero-shot Common Sense Reasoning tasks on \textbf{Mistral-7B-v0.3}.}
\vspace{-2.8em}
\label{tab:appendix_mistral}
\setlength{\tabcolsep}{1mm}
{\resizebox{0.9\textwidth}{!}{
\begin{tabular}{c|l|ccccccccc|c}
& & & & & & & & & & &\\
& & & & & & & & & & &\\
& & & & & & & & & & &\\
\noalign{\vspace{0.2em}}\hline\noalign{\vspace{0.1em}}
\noalign{\vspace{0.1em}}\hline\noalign{\vspace{0.2em}}
\textbf{\#Bits} & \multirow{2}{*}{\textbf{Method}} & \textbf{ARC-e} & \textbf{ARC-c} & \textbf{BoolQ} & \textbf{PIQA} & \textbf{SIQA} & \textbf{HellaS.} & \textbf{OBQA} & \textbf{WinoG.} & \textbf{Avg.}  & \textbf{Wiki2} \\ 
W-A-KV & & ($\uparrow$) & ($\uparrow$) & ($\uparrow$) & ($\uparrow$) & ($\uparrow$) & ($\uparrow$) & ($\uparrow$) & ($\uparrow$) & ($\uparrow$) & ($\downarrow$) \\
\noalign{\vspace{0.2em}}\hline\noalign{\vspace{0.2em}} 
16-16-16 & Full Precision & 81.0 & 57.9 & 84.2 & 82.1 & 48.2 & 80.8 & 59.6 & 73.8 & 71.0 & 5.4 \\
\noalign{\vspace{0.2em}}\hdashline\noalign{\vspace{0.2em}}
\multirow{3}{*}{4-8-16} & RTN & 53.4 & 49.0 & 78.4 & 67.6 & 45.6 & 59.7 & 54.3 & 66.3 & 59.3 & 6.8 \\
 & GPTQ & 38.4 & 41.4 & 74.7 & 59.8 & 42.3 & 45.5 & 50.6 & 61.1 & 51.7 & 8.6 \\
\rowcolor{gray!20} \cellcolor{white} & $\ourse{}$ & 75.4 & 55.7 & 81.9 & 80.3 & 48.2 & 78.1 & 57.8 & 72.7 & 68.8 & 5.7 \\
\rowcolor{gray!20} \cellcolor{white} & $\oursh{}$ & 78.9 & 55.9 & 82.7 & 81.9 & 48.5 & 80.0 & 58.4 & 72.7 & 69.9 & 5.5 \\
\noalign{\vspace{0.2em}}\hdashline\noalign{\vspace{0.2em}}
\multirow{3}{*}{4-8-8} & RTN & 52.9 & 48.7 & 78.5 & 67.3 & 45.5 & 59.4 & 52.7 & 66.4 & 58.9 & 6.7 \\
 & GPTQ & 38.7 & 40.6 & 74.8 & 58.9 & 42.5 & 45.8 & 51.0 & 61.3 & 51.7 & 8.6 \\
\rowcolor{gray!20} \cellcolor{white} & $\ourse{}$ & 76.7 & 54.5 & 82.2 & 80.3 & 50.3 & 78.6 & 59.0 & 73.4 & 69.4 & 5.7 \\
\rowcolor{gray!20} \cellcolor{white} & $\oursh{}$ & 80.1 & 56.9 & 83.9 & 81.5 & 48.6 & 79.9 & 57.2 & 73.0 & 70.2 & 5.5 \\
\noalign{\vspace{0.2em}}\hdashline\noalign{\vspace{0.2em}}
\multirow{3}{*}{4-4-16} & RTN & 39.9 & 24.7 & 50.0 & 57.8 & 39.7 & 34.7 & 33.8 & 50.4 & 41.4 & 449.5 \\
 & GPTQ & 39.4 & 27.1 & 43.8 & 57.3 & 38.4 & 35.6 & 31.4 & 50.0 & 40.4 & 260.8 \\
\rowcolor{gray!20} \cellcolor{white} & $\ourse{}$ & 55.2 & 34.6 & 67.9 & 70.8 & 41.9 & 50.8 & 44.7 & 56.0 & 52.7 & 13.4 \\
\rowcolor{gray!20} \cellcolor{white} & $\oursh{}$ & 76.5 & 53.3 & 80.7 & 80.7 & 48.2 & 78.6 & 57.8 & 71.2 & 68.4 & 5.7 \\
\noalign{\vspace{0.2em}}\hdashline\noalign{\vspace{0.2em}}
\multirow{3}{*}{4-4-4} & RTN & 39.9 & 26.7 & 51.2 & 58.1 & 40.3 & 34.4 & 28.7 & 51.7 & 41.4 & 443.5 \\
 & GPTQ & 40.4 & 28.5 & 43.6 & 57.4 & 39.2 & 35.2 & 33.8 & 52.1 & 41.3 & 249.9 \\
\rowcolor{gray!20} \cellcolor{white} & $\ourse{}$ & 55.4 & 33.3 & 68.5 & 71.4 & 42.4 & 50.9 & 41.0 & 56.3 & 52.4 & 13.7 \\
\rowcolor{gray!20} \cellcolor{white} & $\oursh{}$ & 77.3 & 52.5 & 80.2 & 80.3 & 48.9 & 79.2 & 58.4 & 72.3 & 68.6 & 5.8 \\
\noalign{\vspace{0.2em}}\hline\noalign{\vspace{0.2em}}
\hline
\end{tabular}}}
\end{table*}

\subsection{Results on 3-bit weight quantization}
\begin{table*}[t!]
\renewcommand\arraystretch{0.8}
\centering
\caption{\small 3-bit weight 8-bit activation quantization results on WikiText2 and Zero-shot Common
Sense Reasoning tasks.}
\vspace{-2.8em}
\label{tab:w3a8}
\setlength{\tabcolsep}{1mm}
{\resizebox{0.9\textwidth}{!}{
\begin{tabular}{c|l|ccccccccc|c}
& & & & & & & & & & &\\
& & & & & & & & & & &\\
& & & & & & & & & & &\\
\noalign{\vspace{0.2em}}\hline\noalign{\vspace{0.1em}}
\noalign{\vspace{0.1em}}\hline\noalign{\vspace{0.2em}}
\textbf{\#Bits} & \multirow{2}{*}{\textbf{Method}} & \textbf{ARC-e} & \textbf{ARC-c} & \textbf{BoolQ} & \textbf{PIQA} & \textbf{SIQA} & \textbf{HellaS.} & \textbf{OBQA} & \textbf{WinoG.} & \textbf{Avg.}  & \textbf{Wiki2} \\ 
W-A-KV & & ($\uparrow$) & ($\uparrow$) & ($\uparrow$) & ($\uparrow$) & ($\uparrow$) & ($\uparrow$) & ($\uparrow$) & ($\uparrow$) & ($\uparrow$) & ($\downarrow$) \\
\noalign{\vspace{0.2em}}\hline\noalign{\vspace{0.2em}} 
\multirow{6}{*}{LLaMA-2 7B} & \textit{Full Precision} & 75.0 & 50.8 & 77.3 & 78.9 & 48.5 & 76.0 & 59.3 & 69.5 & 66.9 & 5.5 \\
\noalign{\vspace{0.1em}}\cdashline{2-12}\noalign{\vspace{0.1em}} 
 & RTN & 31.3 & 22.6 & 39.6 & 54.6 & 38.0 & 28.1 & 29.3 & 49.8 & 36.7 & 955.1 \\
 & SmoothQuant & 26.4 & 26.5 & 39.2 & 48.8 & 39.4 & 26.0 & 25.8 & 49.2 & 35.1 & 275,935.2 \\
 & LLM-QAT & 44.0 & 29.5 & 64.4 & 63.3 & 42.2 & 52.7 & 32.6 & 52.3 & 47.6 & 15.2 \\
 & GPTQ & 63.8 & 40.2 & 67.3 & 73.1 & 43.3 & 63.5 & 46.9 & 65.5 & 57.9 & 14.6 \\
\rowcolor{gray!20} \cellcolor{white} & $\oursh{}$ & 71.9 & 47.5 & 74.6 & 76.4 & 47.0 & 71.2 & 53.4 & 67.9 & 63.7 & 6.2 \\
\noalign{\vspace{0.1em}}\hdashline\noalign{\vspace{0.1em}} 
\multirow{6}{*}{LLaMA-2 13B} & \textit{Full Precision} & 75.3 & 51.4 & 79.8 & 80.4 & 50.5 & 79.8 & 56.8 & 72.5 & 68.3 & 5.0 \\
\noalign{\vspace{0.1em}}\cdashline{2-12}\noalign{\vspace{0.1em}} 
 & RTN & 30.4 & 24.6 & 48.8 & 53.8 & 39.8 & 29.0 & 25.4 & 49.5 & 37.6 & 167.8 \\
 & SmoothQuant & 26.1 & 25.5 & 37.8 & 49.0 & 39.4 & 26.1 & 26.4 & 49.5 & 35.0 & 8,979.3 \\
 & LLM-QAT & 27.5 & 20.7 & 40.1 & 51.1 & 38.2 & 26.4 & 27.9 & 50.7 & 35.3 & 256.6 \\
 & GPTQ & 56.5 & 34.5 & 63.3 & 68.9 & 44.2 & 46.0 & 39.8 & 56.3 & 51.2 & 10.8 \\
\rowcolor{gray!20} \cellcolor{white} & $\oursh{}$ & 75.9 & 52.4 & 76.6 & 78.4 & 49.3 & 74.6 & 56.2 & 70.6 & 66.7 & 5.4 \\
\noalign{\vspace{0.1em}}\hdashline\noalign{\vspace{0.1em}} 
\multirow{6}{*}{LLaMA-2 70B} & \textit{Full Precision} & 80.2 & 60.5 & 85.1 & 82.8 & 50.8 & 84.3 & 59.0 & 80.6 & 72.9 & 3.3 \\
\noalign{\vspace{0.1em}}\cdashline{2-12}\noalign{\vspace{0.1em}} 
 & RTN & 51.5 & 30.0 & 59.5 & 65.5 & 40.8 & 40.3 & 31.2 & 51.4 & 46.3 & 66.2 \\
 & SmoothQuant & 26.9 & 22.7 & 38.4 & 49.0 & 38.6 & 25.6 & 25.2 & 52.0 & 34.8 & 6,682.0 \\
 & GPTQ & 72.5 & 49.3 & 72.1 & 76.7 & 46.3 & 69.9 & 51.8 & 72.2 & 63.9 & 9.0 \\
\rowcolor{gray!20} \cellcolor{white} & $\oursh{}$ & 79.4 & 58.7 & 84.4 & 81.6 & 50.5 & 82.3 & 58.3 & 78.6 & 71.7 & 3.8 \\
\noalign{\vspace{0.1em}}\hline\noalign{\vspace{0.1em}} 
\multirow{6}{*}{LLaMA-3 1B} & \textit{Full Precision} & 65.2 & 38.7 & 69.5 & 75.3 & 44.8 & 60.7 & 40.2 & 60.9 & 56.9 & 13.4 \\
\noalign{\vspace{0.1em}}\cdashline{2-12}\noalign{\vspace{0.1em}} 
 & RTN & 32.6 & 28.0 & 54.8 & 55.7 & 39.1 & 34.2 & 29.7 & 47.8 & 40.2 & 2,097.6 \\
 & SmoothQuant & 28.8 & 24.0 & 40.4 & 51.6 & 37.8 & 25.9 & 28.2 & 48.0 & 35.6 & 58,367.5 \\
 & LLM-QAT & 47.0 & 30.4 & 60.3 & 62.8 & 41.6 & 39.9 & 33.6 & 51.8 & 45.9 & 46.9 \\
 & GPTQ & 41.5 & 30.4 & 61.4 & 62.3 & 39.9 & 41.7 & 33.0 & 50.6 & 45.1 & 90.8 \\
\rowcolor{gray!20} \cellcolor{white} & $\oursh{}$ & 58.8 & 36.4 & 63.7 & 68.7 & 44.2 & 51.5 & 38.1 & 56.5 & 52.2 & 17.2 \\
\noalign{\vspace{0.1em}}\hdashline\noalign{\vspace{0.1em}} 
\multirow{6}{*}{LLaMA-3 3B} & \textit{Full Precision} & 68.9 & 47.6 & 79.0 & 76.0 & 52.1 & 71.0 & 50.2 & 66.6 & 63.9 & 10.7 \\
\noalign{\vspace{0.1em}}\cdashline{2-12}\noalign{\vspace{0.1em}} 
 & RTN & 40.1 & 29.5 & 48.8 & 59.3 & 41.6 & 46.0 & 34.4 & 53.4 & 44.1 & 1,178.9 \\
 & SmoothQuant & 27.8 & 21.6 & 38.4 & 50.2 & 38.0 & 25.4 & 26.0 & 50.4 & 34.7 & 17,409.2 \\
 & LLM-QAT & 32.1 & 29.4 & 55.7 & 53.3 & 39.7 & 41.9 & 29.5 & 50.4 & 41.5 & 26.2 \\
 & GPTQ & 48.4 & 33.0 & 65.5 & 63.6 & 41.7 & 57.8 & 38.7 & 57.8 & 50.8 & 176.3 \\
\rowcolor{gray!20} \cellcolor{white} & $\oursh{}$ & 61.8 & 41.4 & 78.2 & 73.0 & 47.4 & 63.3 & 41.0 & 62.8 & 58.6 & 13.7 \\
\noalign{\vspace{0.1em}}\hdashline\noalign{\vspace{0.1em}} 
\multirow{6}{*}{LLaMA-3 8B} & \textit{Full Precision} & 77.6 & 57.7 & 83.3 & 80.7 & 48.7 & 79.6 & 55.8 & 73.7 & 69.6 & 6.1 \\
\noalign{\vspace{0.1em}}\cdashline{2-12}\noalign{\vspace{0.1em}} 
 & RTN & 40.9 & 25.3 & 62.3 & 58.8 & 39.7 & 35.1 & 31.4 & 54.1 & 43.5 & 196.2 \\
 & SmoothQuant & 27.4 & 24.9 & 38.3 & 50.9 & 37.9 & 25.7 & 29.8 & 49.8 & 35.6 & 179,664.5 \\
 & LLM-QAT & 35.9 & 28.0 & 54.3 & 58.5 & 39.8 & 31.7 & 27.7 & 50.9 & 40.8 & 14.9 \\
 & GPTQ & 50.8 & 34.5 & 65.6 & 64.0 & 42.4 & 55.1 & 37.3 & 61.5 & 51.4 & 9.4 \\
\rowcolor{gray!20} \cellcolor{white} & $\oursh{}$ & 74.5 & 50.3 & 79.6 & 77.2 & 46.8 & 74.5 & 50.6 & 70.9 & 65.5 & 7.5 \\
\noalign{\vspace{0.1em}}\hline\noalign{\vspace{0.1em}} 
\multirow{4}{*}{Mistral 7B} & \textit{Full Precision} & 81.0 & 57.9 & 84.2 & 82.1 & 48.2 & 80.8 & 59.6 & 73.8 & 71.0 & 5.4 \\
\noalign{\vspace{0.1em}}\cdashline{2-12}\noalign{\vspace{0.1em}} 
 & RTN & 28.2 & 28.1 & 62.2 & 53.1 & 38.7 & 28.0 & 35.9 & 48.3 & 40.3 & 167.1 \\
 & GPTQ & 31.9 & 32.7 & 63.8 & 54.8 & 40.0 & 31.0 & 36.9 & 52.2 & 42.9 & 29.3 \\
\rowcolor{gray!20} \cellcolor{white} & $\oursh{}$ & 77.7 & 54.1 & 82.2 & 79.9 & 47.7 & 77.5 & 59.4 & 73.8 & 69.0 & 5.8 \\
\noalign{\vspace{0.2em}}\hline\noalign{\vspace{0.2em}}
\hline
\end{tabular}}}
\end{table*}
We present the 3-bit weight and 8-bit activation quantization results across seven models in Table~\ref{tab:w3a8}. Our method, \ours{}, successfully reduces the gap to the full-precision network from the previous $9.0-28.0$ points to $1.2-5.3$ points, demonstrating its effectiveness for low-bit quantization.

\subsection{\textit{Cayley} optimization choice}
\begin{table}[t!]
\renewcommand\arraystretch{0.8}
\centering
\caption{\small Ablation study on Number of training samples and iterations in \textit{Cayley SGD} optimization, using LLaMA-2 7B.}
\vspace{-1.6em}
\label{tab:cayley_sample}
\setlength{\tabcolsep}{1.5mm}
{\resizebox{0.9\textwidth}{!}{
\begin{tabular}{c|c|cc||cccccc}
& & & & & & & & & \\
\noalign{\vspace{0.1em}}\hline\noalign{\vspace{0.1em}}
\hline\noalign{\vspace{0.1em}}
\textbf{\#Bits} & \multirow{2}{*}{Task} & \multicolumn{2}{c||}{\# Training sample} & \multicolumn{5}{c}{\# Training iterations} \\
\noalign{\vspace{0.1em}}
(W-A-KV) &  & 128 & 800 & 10 & 25 & 50 & 100 & 200\\
\noalign{\vspace{0.1em}}\hline\noalign{\vspace{0.1em}}
4-4-4 & Wiki ($\downarrow$) & 6.2 $_{\pm 0.03}$ & 6.2 $_{\pm 0.03}$ & 6.6 $_{\pm 0.02}$ & 6.4 $_{\pm 0.02}$ & 6.3 $_{\pm 0.03}$ & 6.2 $_{\pm 0.03}$ & 6.2 $_{\pm 0.05}$\\
\noalign{\vspace{0.1em}}\hline\noalign{\vspace{0.1em}}
\hline
\end{tabular}}}
\vspace{2em}
\end{table}

In Table~\ref{tab:cayley_sample}, we evaluate the impact of varying the number of samples and iterations used in \textit{Cayley} optimization. Given the limited trainable parameters in the rotation matrix and its constraint optimization nature, minimal calibration data and iterations are sufficient to optimize the rotation for better quantization. The findings indicate that rotation optimization is resilient to modifications in the number of samples. Even though we used 800 samples in our experiments, reducing this to 128 samples does not lead to a significant change in the perplexity. Furthermore, we examined the optimal number of iterations and found that the wiki perplexity ceases to decrease and stabilizes at 100 iterations. Consequently, we chose to use 100 iterations in all our experiments.

\subsection{Quantization choice} 
\begin{table}[t!]
\renewcommand\arraystretch{0.6}
\centering
\caption{\small Ablation of symmetric and asymmetric quantization and range clipping options on LLaMA-2 7B. }
\vspace{-1em}
\label{tab:Q_choice}
\setlength{\tabcolsep}{1.5mm}
{\resizebox{0.9\textwidth}{!}{
\begin{tabular}{c|cc|cc|cc|cccccc}
\noalign{\vspace{0.2em}}\hline\noalign{\vspace{0.1em}}
\hline\noalign{\vspace{0.2em}}
\textbf{\#Bits} & & & & & \multicolumn{2}{c}{\textbf{RTN}} &\multicolumn{2}{c}{\textbf{GPTQ}} \\
(W-A-KV) & K asym  & K clip & A asym & A clip & Zero-shot Avg. ($\uparrow$)  & Wiki ($\downarrow$) & Zero-shot Avg. ($\uparrow$)  & Wiki ($\downarrow$) \\
\noalign{\vspace{0.2em}}\hline\noalign{\vspace{0.2em}}
4-4-16 & -- & -- & \xmark & \xmark &         61.2 $_{\pm 0.6}$ & 6.3 & 63.3 $_{\pm 0.4}$ & 6.0 \\
4-4-16 & -- & -- & \cmark & \xmark &         61.8 $_{\pm 0.4}$ & 6.1 & 64.0 $_{\pm 0.5}$ & 5.9 \\
4-4-16 & -- & -- & \cmark & \cmark &         62.1 $_{\pm 0.6}$ & 6.0 & 64.0 $_{\pm 0.4}$ & 5.9 \\
\noalign{\vspace{0.2em}}\hline\noalign{\vspace{0.2em}}
4-4-4 & \xmark & \xmark & \cmark & \xmark &  61.4 $_{\pm 0.5}$ & 6.2 & 63.7 $_{\pm 0.4}$ & 6.0 \\
4-4-4 & \cmark & \xmark & \cmark & \xmark &  61.5 $_{\pm 0.6}$ & 6.2 & 63.7 $_{\pm 0.3}$ & 5.9 \\
4-4-4 & \cmark & \cmark & \cmark & \xmark &  61.5 $_{\pm 0.3}$ & 6.2 & 63.7 $_{\pm 0.2}$ & 5.9 \\
\noalign{\vspace{0.2em}}\hline\noalign{\vspace{0.1em}}
\hline
\end{tabular}}}
\end{table}
We conduct an ablation study on symmetric vs asymmetric quantization and whether to clip the min-max ranges or not during activation and KV-cache quantization. The results in Table~\ref{tab:Q_choice} show that for both activation quantization and KV-cache quantization, asymmetric quantization outperforms symmetric quantization. In the clip settings, we set the activation clipping ratio to 0.9 and the KV-cache clipping ratio to 0.95 as suggested in the previous works~\citep{zhao2023atom}. However, the results show that clipping the range or not does not impact the final result significantly. Therefore we opt for no clipping, \textit{i.e.}, using the min-max quantization for activation and KV cache quantization across our experiments due to its simplicity.

\subsection{Calibration data choice}
\begin{table*}[t!]
\renewcommand\arraystretch{0.8}
\centering
\caption{\small Ablation study on calibration data choice using LLaMA-2 7B.}
\vspace{-2em}
\label{tab:cali_data_ablation}
\setlength{\tabcolsep}{1mm}
{\resizebox{0.9\textwidth}{!}{
\begin{tabular}{l|c|ccccccccc|c}
& & & & & & & & & & &\\
\noalign{\vspace{0.2em}}\hline\noalign{\vspace{0.1em}}
\noalign{\vspace{0.1em}}\hline\noalign{\vspace{0.2em}}
\textbf{Calibration} & \textbf{\#Bits} & \textbf{ARC-e} & \textbf{ARC-c} & \textbf{BoolQ} & \textbf{PIQA} & \textbf{SIQA} & \textbf{HellaS.} & \textbf{OBQA} & \textbf{WinoG.} & \textbf{Avg.}  & \textbf{Wiki2} \\ 
\textbf{Data} & $_\text{(W-A-KV)}$ & ($\uparrow$) & ($\uparrow$) & ($\uparrow$) & ($\uparrow$) & ($\uparrow$) & ($\uparrow$) & ($\uparrow$) & ($\uparrow$) & ($\uparrow$) & ($\downarrow$) \\
\noalign{\vspace{0.2em}}\hline\noalign{\vspace{0.2em}} 
Wiki2 & 4-4-16 & 72.1 & 47.5 & 74.4 & 77.0 & 47.3 & 73.2 & 54.4 & 66.9 & 64.1 & 5.9 \\
Wiki2 & 4-4-4  & 72.6 & 47.5 & 73.9 & 77.0 & 47.2 & 73.0 & 54.1 & 66.9 & 64.0 & 5.9 \\
\noalign{\vspace{0.2em}}\hdashline\noalign{\vspace{0.2em}}
C4 & 4-4-16 & 72.5 & 47.3 & 74.8 & 77.6 & 47.7 & 73.2 & 55.4 & 66.2 & 64.3 & 5.9 \\
C4 & 4-4-4  & 72.5 & 47.9 & 74 & 78.4 & 46.7 & 73.1 & 55.5 & 66.4 & 64.3 & 6 \\
\hline\noalign{\vspace{0.2em}}
\hline
\end{tabular}}}
\end{table*}
To assess the robustness of SpinQuant with respect to calibration data used in rotation optimization we use C4 dataset~\citep{raffel2020c4} as calibration data and performe experiments on the LLaMA-2 7B model. The results in Table~\ref{tab:cali_data_ablation} reflect that using C4 datasets yields consistent results with utilizing the Wiki dataset, showing that SpinQuant is robust to calibration data choice.

\subsection{Latency measurement on GPU}

In light of the available Tensor cores in NVIDIA's Hopper (H100) architecture, we provide the whole network end-to-end speed test result of W-fp8-A-fp8 quantization on H100 GPU, both with and without Hadamard transformations. Specifically, we utilize FP8 GEMM from the FBGEMM repo~\footnote{https://github.com/pytorch/FBGEMM/blob/main/fbgemm\_gpu/experimental/gemm/triton\_gemm/ fp8\_gemm.py}, which incorporates dequantization via epilogue fusion. We also leverage the Tensor Core-based Hadamard transform kernel~\footnote{https://github.com/pytorch-labs/applied-ai/blob/main/kernels/cuda/inference/hadamard\_transform/ hadamard\_transform.cpp} to minimize the overhead of the online Hadamard transform. The end-to-end speed test results of LLaMA-3 70B are detailed in Table~\ref{tab:7_gpu_tensor_core_speed}. When implemented meticulously, SpinQuant with Hadamard rotation sees marginal difference in the latency compared to without Hadamard rotation.

\begin{table}[t]
\renewcommand\arraystretch{0.6}
\centering
\caption{\small Real-time end-to-end speed measurement of LLaMA-3 70B on NVIDIA's Hopper (H100) GPU. TTFT (Time to First Token) and TTIT (Time to Iterative Token) are performance metrics to measure the pre-filling speed and decoding speed, respectively.}
\vspace{-1em}
\label{tab:7_gpu_tensor_core_speed}
\setlength{\tabcolsep}{1.5mm}
{\resizebox{0.64\textwidth}{!}{
\begin{tabular}{lcccccccccc}
\hline\noalign{\vspace{0.2em}}
\hline\noalign{\vspace{0.2em}}
& \multicolumn{2}{c}{SpinQuant without hadamard} & \multicolumn{2}{c}{SpinQuant with hadamard} \\
\hline\noalign{\vspace{0.2em}}
 & TTFT (ms) & TTIT (ms) & TTFT (ms) & TTIT (ms) \\
BS=1 T=4096 & 153.58 & 9.85 & 158.25 & 10.15 \\
BS=8 T=4096 & 1205.47 & 10.6 & 1243.48 & 10.94 \\
BS=32 T=4096 & 5008.25 & 13.83 & 5147.59 & 14.2 \\
\hline\noalign{\vspace{0.2em}}
\hline\noalign{\vspace{0.2em}}
\end{tabular}}}
\end{table}

\subsection{Optimization time}
In Table~\ref{tab:1_optimization_time}, we show the comparison of optimization time between GPTQ and SpinQuant. SpinQuant requires a scale of optimization time similar to that of GPTQ. The additional optimization time required by SpinQuant is worthwhile considering the substantial improvements it offers over GPTQ.
\begin{table}[t]
\renewcommand\arraystretch{0.6}
\centering
\caption{\small Optimization time comparison between GPTQ and SpinQuant.}
\vspace{-1em}
\label{tab:1_optimization_time}
\setlength{\tabcolsep}{1.5mm}
{\resizebox{0.7\textwidth}{!}{
\begin{tabular}{lcccccc}
\hline\noalign{\vspace{0.2em}}
\hline\noalign{\vspace{0.2em}}
& llama3 1B & llama3 3B & llama3 8B & llama2 7B & llama2 13B & Mistral \\
\noalign{\vspace{0.2em}}\hline\noalign{\vspace{0.2em}}
SpinQuant & 13 min & 18 min & 30 min & 25 min & 30 min & 16 min \\
GPTQ & 8 min & 13 min & 20 min & 18 min & 25 min & 12 min \\  
\hline\noalign{\vspace{0.2em}}
\hline\noalign{\vspace{0.2em}}
\end{tabular}}}
\end{table}

\subsection{Ablation study on RTN vs GPTQ}
SpinQuant is fully compatible with both GPTQ and naive RTN. To isolate the contributions of GPTQ and rotation to overall performance, we present results for SpinQuant combined with RTN in the W4A4KV16 quantization scenario in Table~\ref{tab:2_RTN_vs_GPTQ}. Our analysis indicates that the primary accuracy gains are attributed to the incorporation of learned rotations, which enhances accuracy by 6.5 $\sim$ 20.9 percentage points over previous methods (including GPTQ). The subsequent integration of GPTQ further boosts performance by up to 2.3 percentage points.
\begin{table}[t]
\renewcommand\arraystretch{0.6}
\centering
\caption{\small Ablation study on SpinQuant combined with RTN or GPTQ in the W4A4KV16 quantization scenario.}
\vspace{-1em}
\label{tab:2_RTN_vs_GPTQ}
\setlength{\tabcolsep}{1.5mm}
{\resizebox{0.8\textwidth}{!}{
\begin{tabular}{lcccccccccc}
\hline\noalign{\vspace{0.2em}}
\hline\noalign{\vspace{0.2em}}
 & \multicolumn{2}{c}{LLaMA-3 8B} & \multicolumn{2}{c}{LLaMA-2 7B} & \multicolumn{2}{c}{LLaMA-2 13B} & \multicolumn{2}{c}{LLaMA-2 70B}  \\
Method & Zero-shot & Wiki & Zero-shot & Wiki & Zero-shot & Wiki & Zero-shot & Wiki \\
\hline\noalign{\vspace{0.2em}}
Full-precision & 69.6 & 6.1 & 66.9 & 5.5 & 68.3 & 5.0 & 72.9 & 3.3\\
\hdashline\noalign{\vspace{0.2em}}
RTN & 43.9 & 2e2 & 35.6 & 2e3 & 35.3 & 7e3 & 35.1 & 2e5 \\
SmoothQuant & 40.3 & 9e2 & 41.8 & 3e2 & 44.9 & 34.5 & 57.7 & 57.1 \\
LLM-QAT & 44.9 & 42.9 & 47.8 & 12.9 & 34.3 & 4e3 & -- & -- \\
GPTQ & 40.6 & 2e2 & 36.8 & 9e3 & 35.2 & 5e3 & 35.5 & 2e6 \\
$\oursh$ (RTN) & 64.6 & 7.7 & 61.8 & 6.1 & 65.8 & 5.4 & 71.1 & 3.9 \\
$\oursh$ (GPTQ) & \textbf{65.8} & \textbf{7.1} & \textbf{64.1} & \textbf{5.9} & \textbf{67.2} & \textbf{5.2} & \textbf{71.0} & \textbf{3.8} \\ 
\hline\noalign{\vspace{0.2em}}
\hline\noalign{\vspace{0.2em}}
\end{tabular}}}
\end{table}

\subsection{Weight-only quantization}
We also include a comparison of SpinQuant performance under weight-only quantization in Table~\ref{tab:4_weight_only}. The weight-only results show that SpinQuant consistently achieves higher accuracy than AWQ and other previous work.
\begin{table}[t]
\renewcommand\arraystretch{0.6}
\centering
\caption{\small A comparison of SpinQuant performance under 4-bit weight-only quantization. }
\vspace{-1em}
\label{tab:4_weight_only}
\setlength{\tabcolsep}{1.5mm}
{\resizebox{0.9\textwidth}{!}{
\begin{tabular}{lccccccc}
\hline\noalign{\vspace{0.2em}}
\hline\noalign{\vspace{0.2em}}
& \multicolumn{2}{c}{\textbf{LLaMA-2 7B}} & \multicolumn{2}{c}{\textbf{LLaMA-2 13B}}& \multicolumn{2}{c}{\textbf{LLaMA-2 70B}} \\
\textbf{Method} & Zero-shot Avg. ($\uparrow$) & Wiki ppl($\downarrow$) & Zero-shot Avg.($\uparrow$) & Wiki ppl($\downarrow$) & Zero-shot Avg.($\uparrow$) & Wiki ppl($\downarrow$) \\
\hline\noalign{\vspace{0.2em}}
Full-precision & 66.9 & 5.5 & 68.3 & 5.0 & 72.9 & 3.3\\
\hdashline\noalign{\vspace{0.2em}}
RTN & 63.6 & 7.2 & 57.9 & 6.4 & 69.2 & 4.6  \\
SmoothQuant & 59.1 & 7.5 & 63.3 & 6.1 & 70.2 & 4.1  \\
GPTQ & 64.5 & 11.3 & 64.7 & 5.6 & 71.9 & 3.9  \\
AWQ & -- & 6.2 & -- & 5.1  &-- & --  \\
\oursh & \textbf{65.9} & \textbf{5.6} & \textbf{68.5} & \textbf{5.0} & \textbf{72.6} & \textbf{3.5}  \\
\hline\noalign{\vspace{0.2em}}
\hline\noalign{\vspace{0.2em}}
\end{tabular}}}
\end{table}

\subsection{Few-shot results on instruction-finetuned models}
\begin{table}[thb!]
\renewcommand\arraystretch{0.6}
\centering
\caption{\small Results of applying SpinQuant to instruction-finetuned LLaMA 3.2 1B and 3B models.}
\vspace{-1em}
\label{tab:5_more_tasks_instruct_model}
\setlength{\tabcolsep}{1.5mm}
{\resizebox{0.8\textwidth}{!}{
\begin{tabular}{lcccccc}
\hline\noalign{\vspace{0.2em}}
\hline\noalign{\vspace{0.2em}}
 & & \multicolumn{2}{c}{\textbf{llama3.2 1B}} & \multicolumn{2}{c}{\textbf{llama3.2 3B}} \\
\hline\noalign{\vspace{0.2em}}
 & BF16 & Vanilla RTN & SpinQuant & BF16 & Vanilla RTN & SpinQuant \\
\hdashline\noalign{\vspace{0.2em}}
MMLU (5-shot) & 49.3 & 43.4 & 47.3 & 63.4 & 60.5 & 62.0 \\
TLDR9+ (test, 1-shot rougeL) & 16.8 & 14.9 & 16.7 & 19.0 & 19.1 & 19.2 \\
\hline\noalign{\vspace{0.2em}}
\hline\noalign{\vspace{0.2em}}
\end{tabular}}}
\end{table}

We further conduct experiments applying SpinQuant to instruction-finetuned LLaMA 3.2 1B and 3B models in Table~\ref{tab:5_more_tasks_instruct_model}. We present the results for few-shot learning scenarios. SpinQuant W4A8 quantized models demonstrate significant improvements in 5-shot accuracy on the MMLU benchmark and 1-shot rouge score on the TLDR9 summarization benchmark. It significantly closed the gap to the BF16 baseline.

\section{Analysis}
\subsection{Gradient Analysis}
On the one hand, we have shown that the class of LLMs we are interested in are rotation invariant, \textit{i.e.} the full-precision model output does not change regardless of what $\rotm$ is. On the other hand, we are claiming that some $\rotm$ are better than others for quantized LLM and that better $\rotm$ can be learned with backpropagation on equation \eqref{eq:optimization_problem}. To reconcile these seemingly conflicting claims, we inspect the gradient of the output of a single linear, $W$, and activations, $X$, which are both rotated and quantized:
\begin{align}\label{eq:rotation_derivative}
     \frac{\partial \sum_{ij} \left(Q(W \rotm^{-1}) Q(\rotm X)) \right)_{ij}}{\partial \rotm_{mn}} &= \sum_{ij} -(W \rotm^{-1})_{im} (\rotm^{-1} Q(\rotm X))_{nj} + Q(W \rotm^{-1})_{im}X_{nj}
\end{align}
We see that equation \eqref{eq:rotation_derivative}:
\begin{itemize}
    \item is non-zero in general, which validates our approach of using backpropagation to learn \rotm
    \item reduces to $0$ when quantization is not present, which validates the claim that it only makes sense to learn $\rotm$ for quantized models
    \item demonstrates that two components move the gradient with respect to \rotm{} away from 0: 1) differences in quantized and unquantized rotated weights; 2) differences in quantized and unquantized rotated activations
\end{itemize}

\subsection{Loss Analysis}
\label{sec:analysis}

\begin{table}[htb!]
\centering
\caption{\small Average end-to-end signal to quantization noise ratio (dB) for LLaMA-2 7B with weights and activations quantized to 4 bits on wiki2 test set}
\vspace{-1em}
\label{tab:quantization_error}
\resizebox{0.4\textwidth}{!}{
\begin{tabular}{c|c|c}
\noalign{\vspace{0.2em}}\hline\noalign{\vspace{0.1em}}
\hline\noalign{\vspace{0.2em}}
\textbf{$\rotm = I$} & Randomly initialized $\rotm$ & Learned $\rotm$ \\ \hline
-2.9 & 0.9 & 6.8 \\ \hline
\end{tabular}}
\vspace{-1em}
\end{table}

\begin{figure}[h]
\centering
\begin{subfigure}{.3\textwidth}
  \centering
  \includegraphics[width=\linewidth]{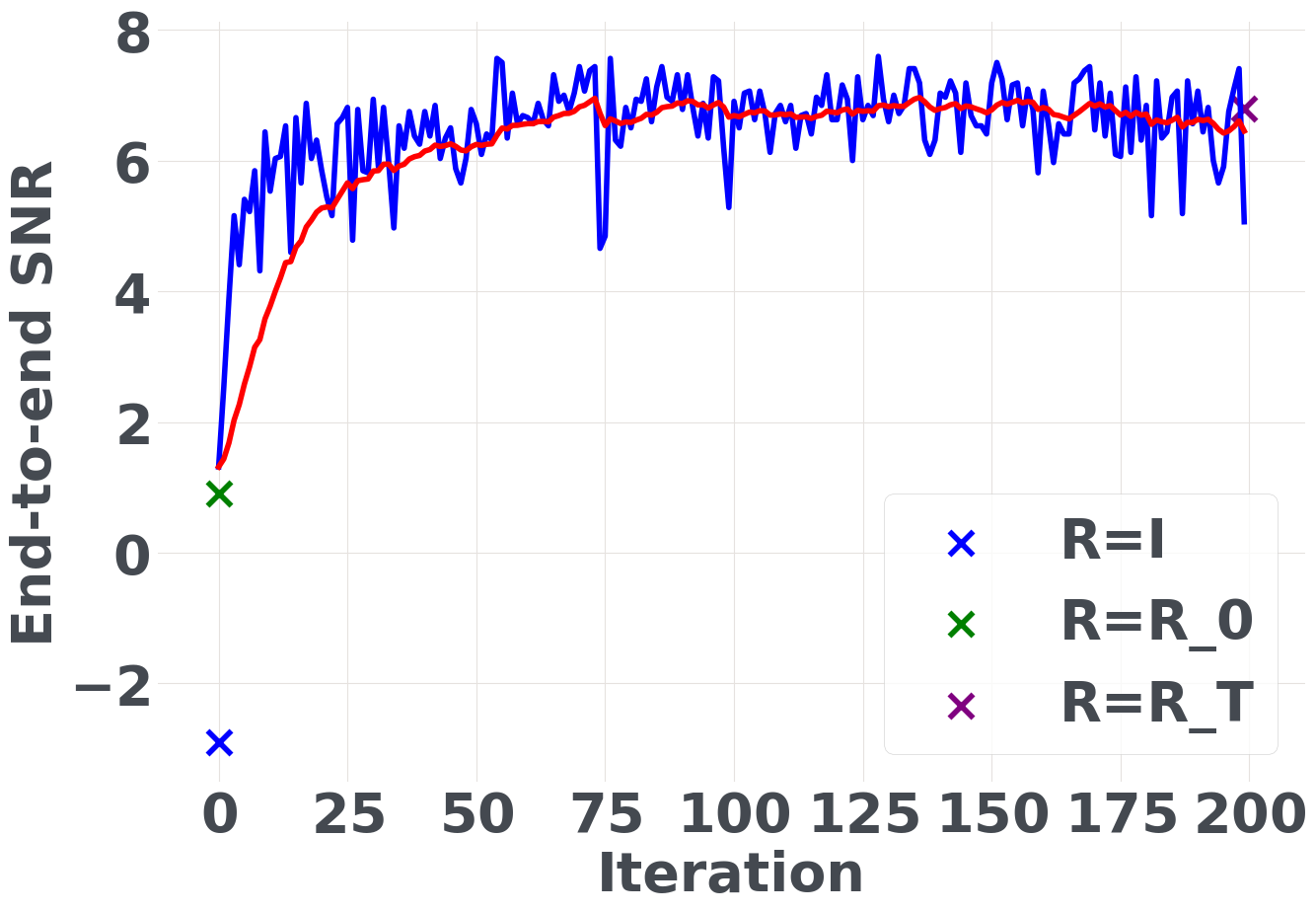}
  \caption{}
  \label{fig:end_to_end_snr_training}
\end{subfigure}
\begin{subfigure}{.3\textwidth}
  \centering
  \includegraphics[width=\linewidth]{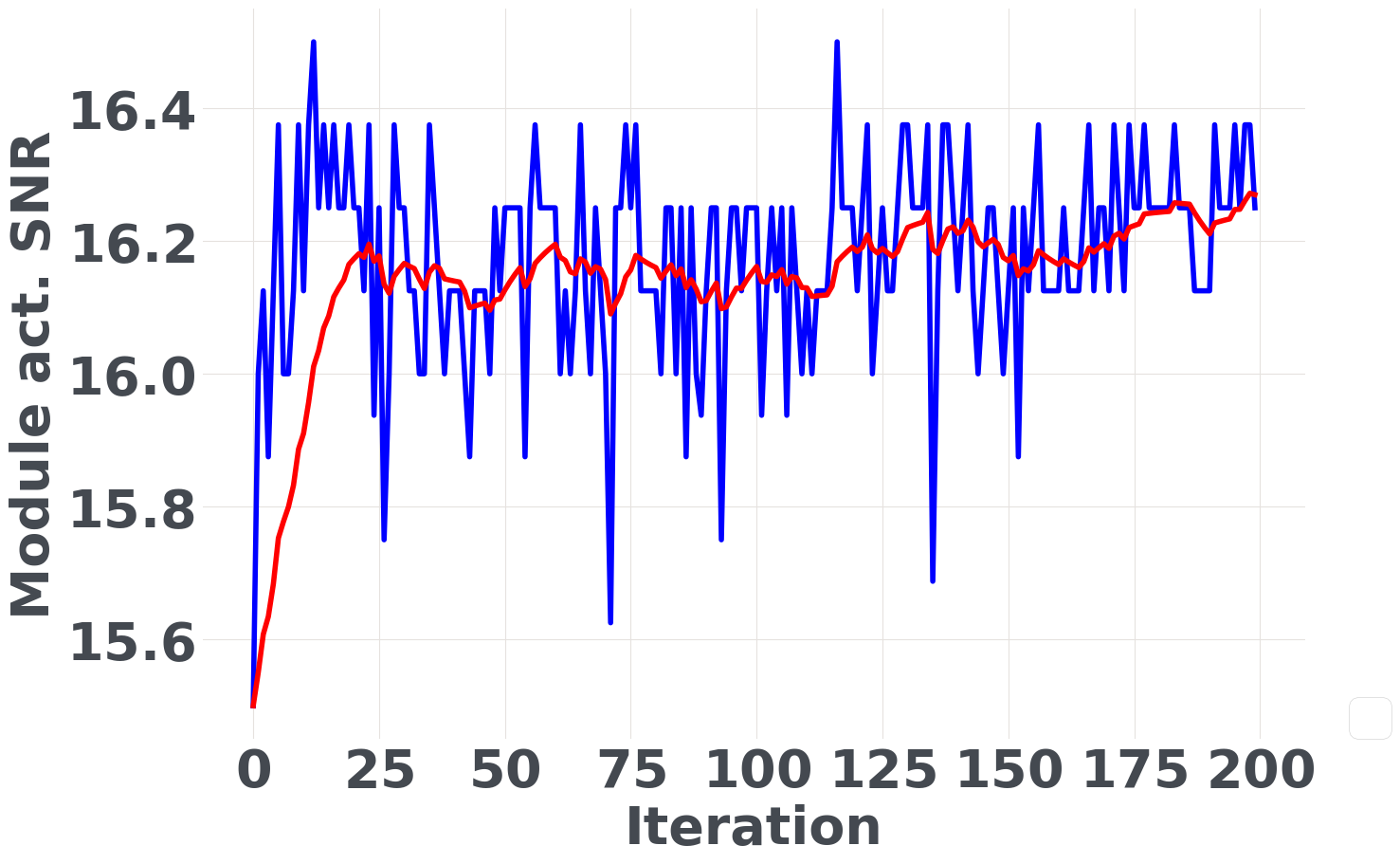}
  \caption{}
  \label{fig:layer_level_snr}
\end{subfigure}
\begin{subfigure}{.3\textwidth}
  \centering
  \includegraphics[width=\linewidth]{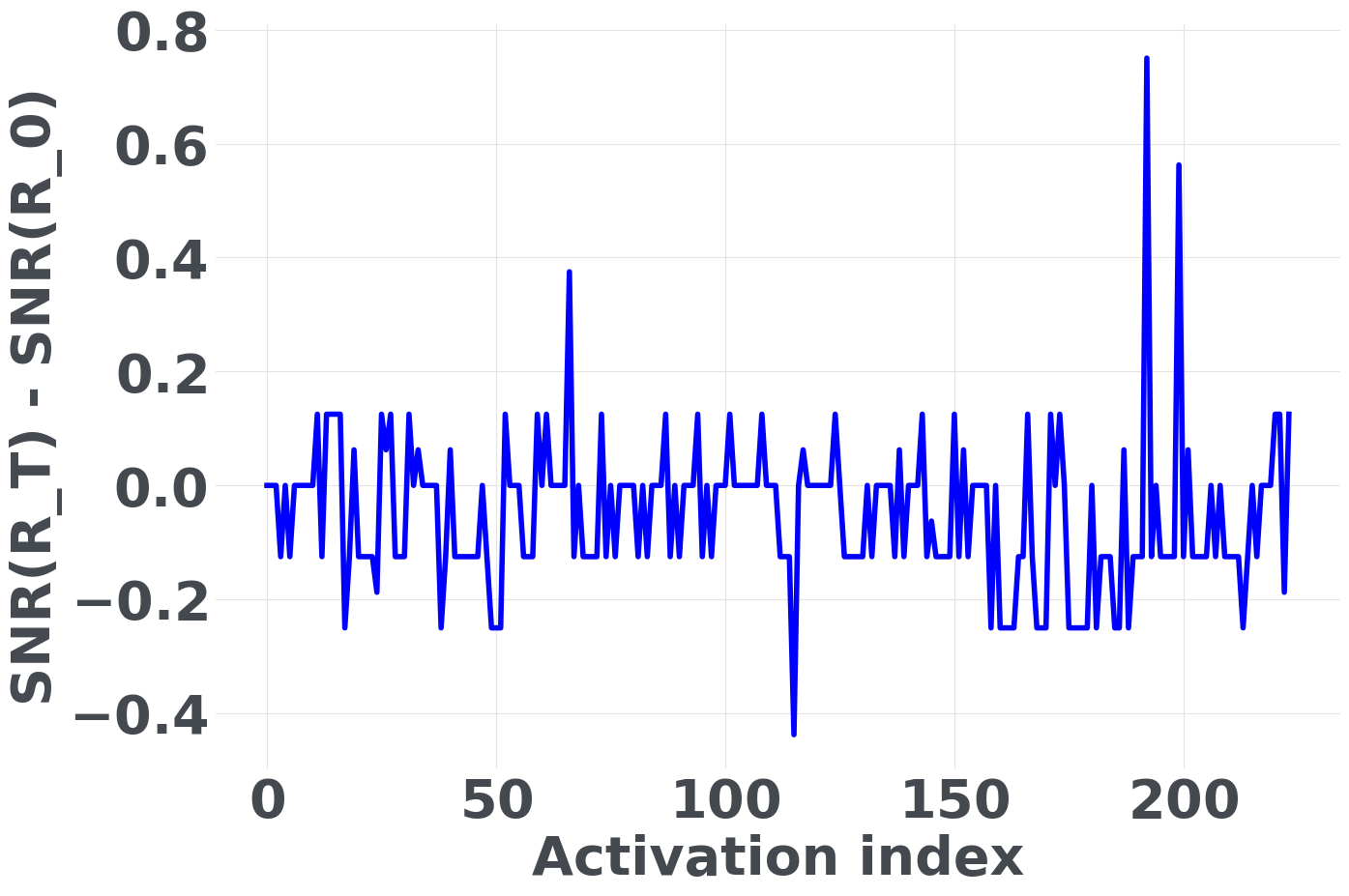}
  \caption{}
  \label{fig:layer_level_improvement}
\end{subfigure}
\caption{\small Training curves for LLaMA-2 7B with 4-bit weights and 4-bit activations in wiki2 train set. (a) End-to-end quantization SNR. $\rotm{}_0$ and $\rotm{}_T$ denote randomly initialized rotation and learned rotation after $T=200$ iterations; (b) Activation quantization. SNR for layer 27 attention out projection; (c) Improvement in activation quantization SNR after optimization of \rotm{} for each layer.}
\label{fig:test}
\end{figure}

While Sec. \ref{sec:experiments} shows that learning $\rotm$ yields significant benefits on zero-shot reasoning tasks, in this section we shed some light on why our method is able to achieve accuracy gains. Intuitively, we expect the end-to-end signal to (quantization) noise ratio (SNR) to improve as a result of learning \rotm{}. In other words, learning \rotm{} should bring the quantized model output closer to the floating point model output. As Table~\ref{tab:quantization_error} shows, we observe an SNR improvement of 3.8 dB when introducing a random \rotm{} into LLaMA-2 7B with weights/activations quantized to 4 bits, and then an additional $5.9$dB improvement after learning \rotm{}, all measured on the WikiText2~\citep{merity2016wiki2} test set. Figure~\ref{fig:end_to_end_snr_training} shows that the batch-level training set SNR during $\rotm$ training progressively improves as expected, as well as the layer-level SNR for a particular layer in Figure~\ref{fig:layer_level_snr}. Digging a bit deeper, Figure~\ref{fig:layer_level_improvement} shows the layer-level SNR improvement for each layer as a result of training \rotm{}. We see that, perhaps counter-intuitively, layer-level SNR improves significantly for a few layers, but does not change much for most layers, and even gets worse for one of the layers. We hypothesize that: 1) certain layers have a disproportionate impact on model output or have a disproportionately low quantization SNR without rotation; 2) The process of optimizing \rotm{} rotates the residual stream basis such as to prioritize improving the SNR of such layers, possibly at the cost of hurting less important layers.

\section{Distribution visualizations before and after rotation}

\begin{figure}[t!]
    \centering
    \includegraphics[width=0.9\linewidth]{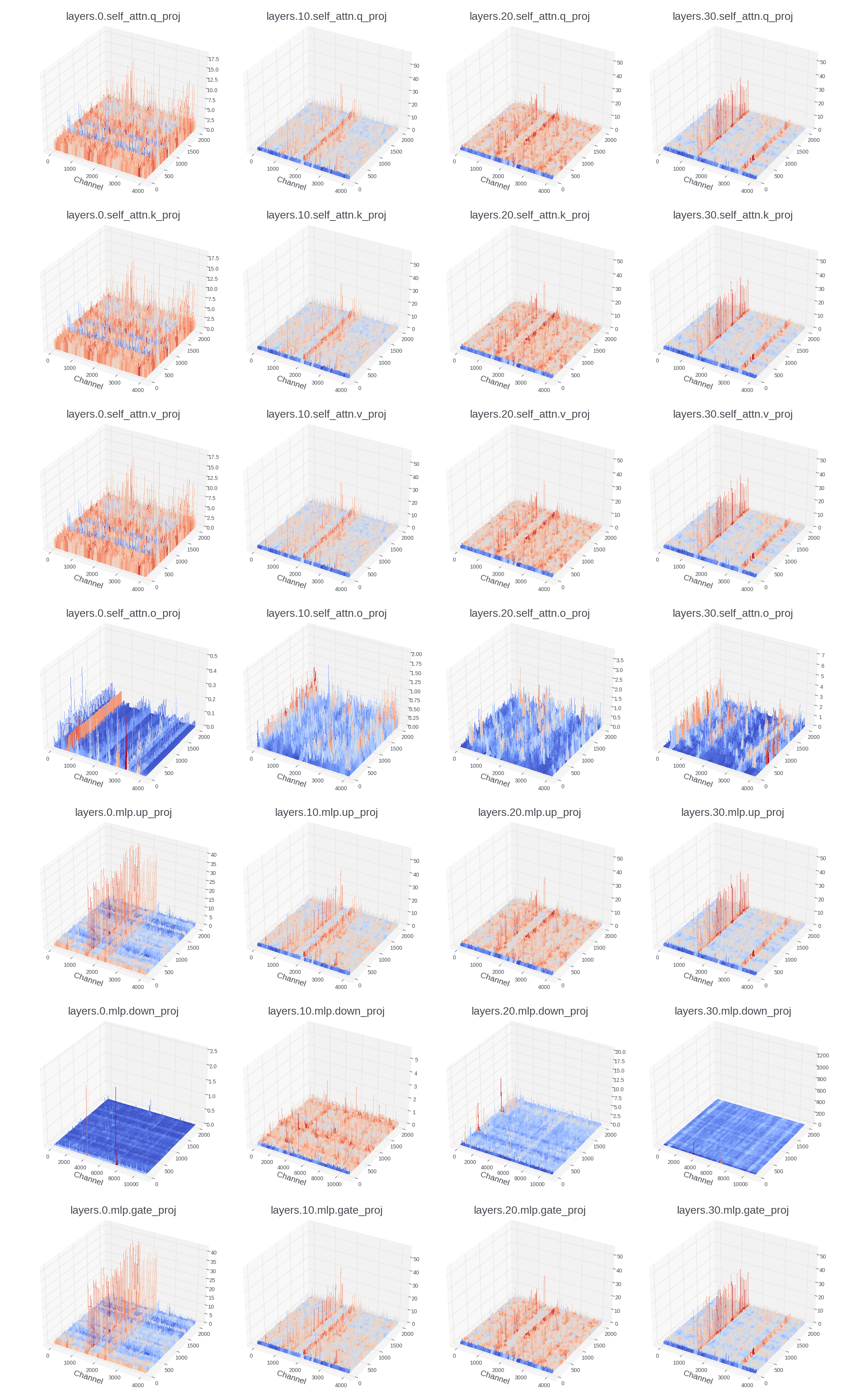}
    \caption{\small Magnitude of the input activations of a linear layer in \{1$^{st}$, 11$^{th}$, 21$^{st}$, and 31$^{st}$\} blocks in LLaMA-2 7B model \textbf{before rotation}.}
    \label{fig:activation_before_rotation}
\end{figure}
\begin{figure}[t!]
    \centering
    \includegraphics[width=0.9\linewidth]{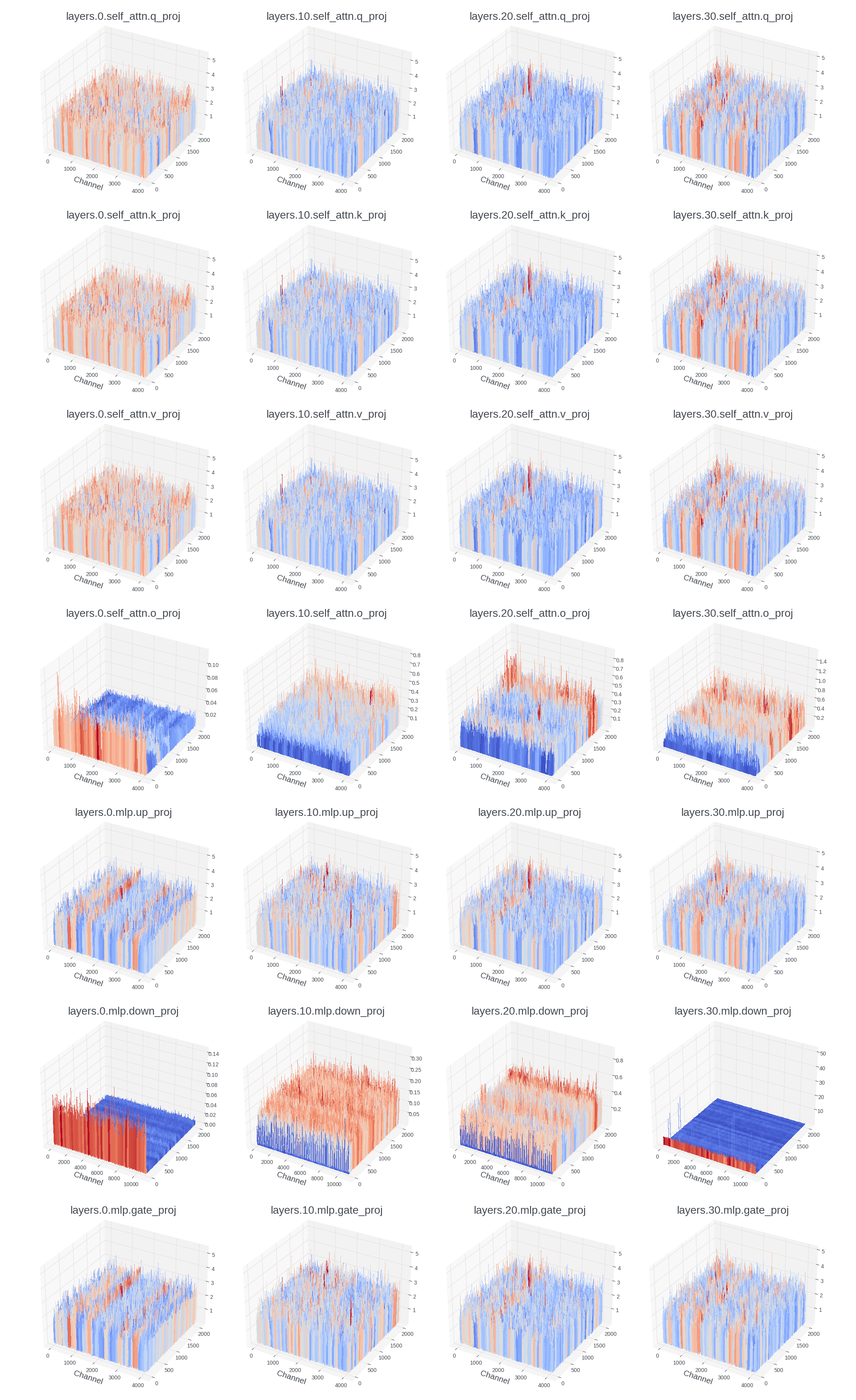}
    \caption{\small Magnitude of the input activations of a linear layer in \{1$^{st}$, 11$^{th}$, 21$^{st}$, and 31$^{st}$\} blocks in LLaMA-2 7B model \textbf{after rotation}.}
    \label{fig:activation_after_rotation}
\end{figure}
\begin{figure}[t!]
    \centering
    \includegraphics[width=0.9\linewidth]{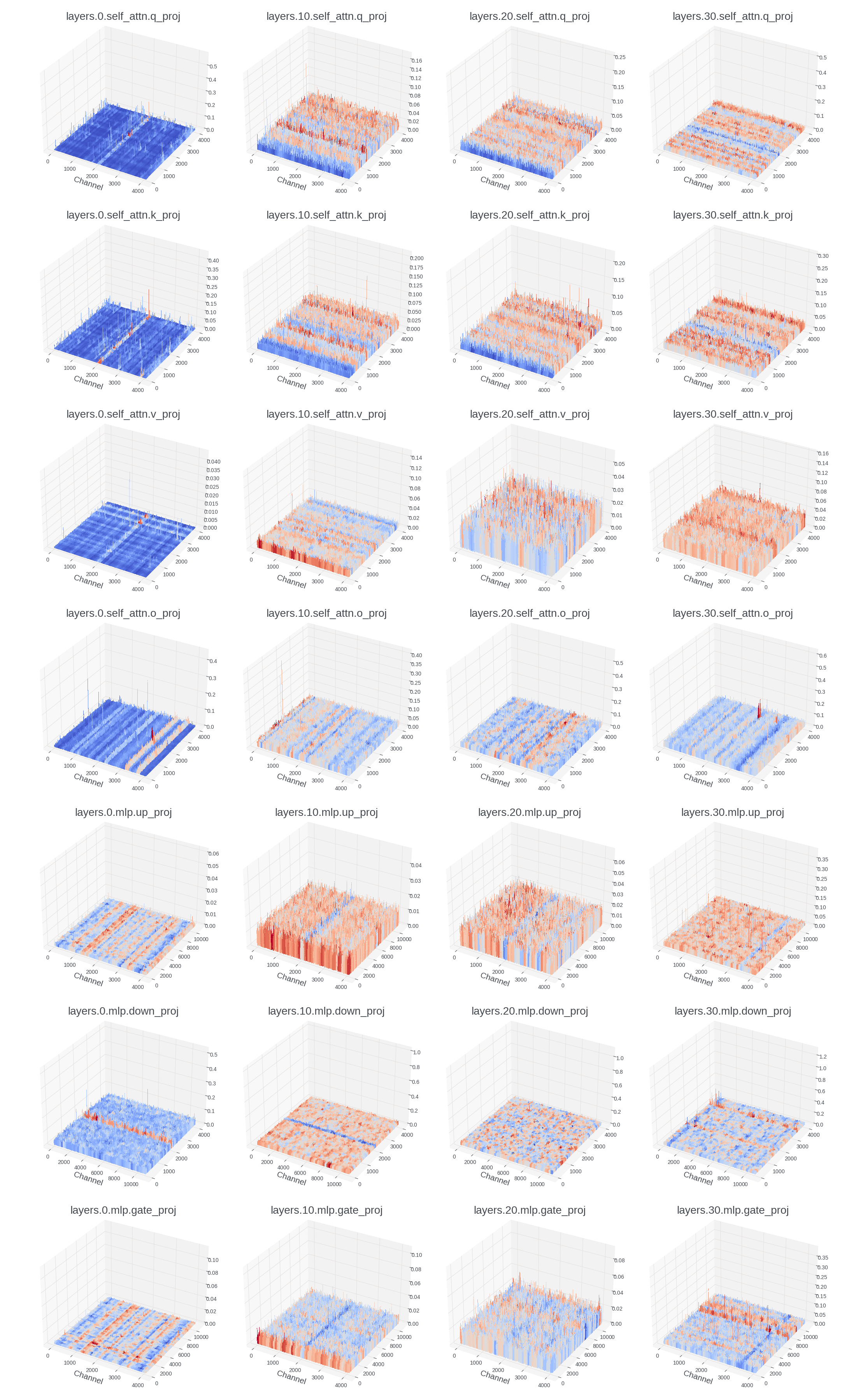}
    \caption{\small Magnitude of the weights of a linear layer in \{1$^{st}$, 11$^{th}$, 21$^{st}$, and 31$^{st}$\} blocks in LLaMA-2 7B \textbf{before rotation}.}
    \label{fig:weight_before_rotation}
\end{figure}
\begin{figure}[t!]
    \centering
    \includegraphics[width=0.9\linewidth]{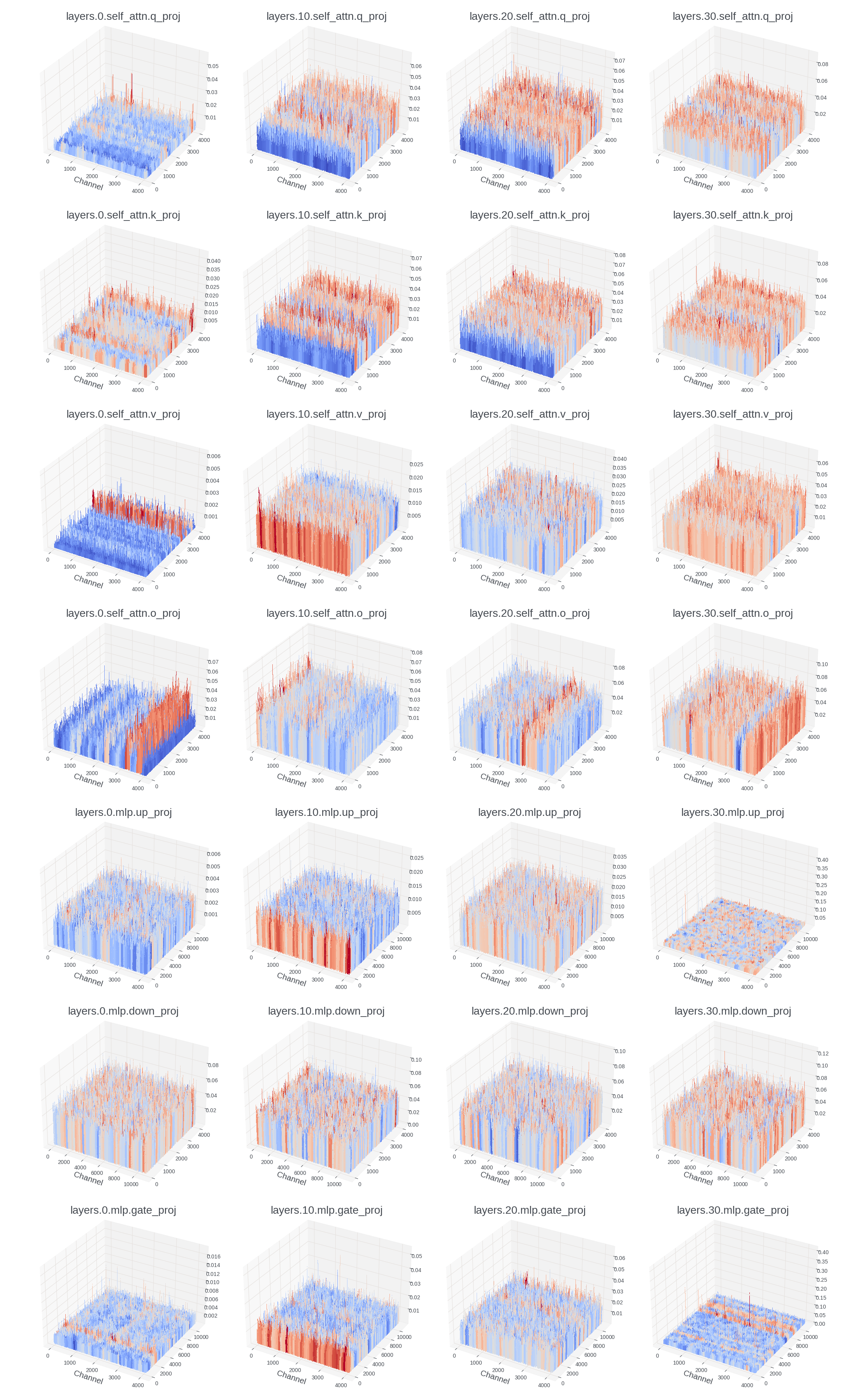}
    \caption{\small Magnitude of the weights of a linear layer in \{1$^{st}$, 11$^{th}$, 21$^{st}$, and 31$^{st}$\} blocks in LLaMA-2 7B \textbf{after rotation}.}
    \label{fig:weight_after_rotation}
\end{figure}
We present visualizations of the activation distributions before and after rotation in Figures~\ref{fig:activation_before_rotation} and \ref{fig:activation_after_rotation}, respectively. Similarly, the weight distributions before and after rotation are depicted in Figures~\ref{fig:weight_before_rotation} and \ref{fig:weight_after_rotation}. Overall, after rotation, the extreme values are attenuated, and the distribution exhibits no noteworthy outliers across the token dimension. Additionally, we make an interesting observation: in several activation layers, the first token displays substantial values in multiple channels. After rotation, this outlier is distributed across all channels of the first token. Although per-token activation quantization can readily manage this distribution, investigating the source of these outliers and reducing them prior to applying $\ours$ might further enhance quantization accuracy, which could be a potential future research direction.

\end{document}